\documentclass[lettersize,journal]{IEEEtran}
\usepackage{amsmath,amsfonts}
\usepackage[ruled,linesnumbered,vlined]{algorithm2e}
\usepackage{lineno} 
\usepackage{array}
\usepackage[caption=false,font=normalsize,labelfont=sf,textfont=sf]{subfig}
\usepackage{textcomp}
\usepackage{hyperref}
\usepackage{color}  
\hypersetup{
    colorlinks=blue,
    linkcolor=blue,
    citecolor=blue,
    filecolor=blue,
    urlcolor=blue
}
\usepackage{multirow}
\usepackage{booktabs}
\usepackage{stfloats}
\usepackage{tabularx}
\usepackage{url}
\usepackage{verbatim}
\usepackage{graphicx}
\usepackage{cite}
\usepackage{multirow}
\usepackage{enumitem}
\usepackage[table]{xcolor}
\usepackage{xcolor}
\usepackage{graphicx}
\usepackage{amsmath}
\usepackage{array}
\usepackage{longtable}
\usepackage{tcolorbox}
\usepackage{fontawesome5}  
\usepackage{tcolorbox}
\usepackage{float}

\definecolor{promptbg}{RGB}{245,247,250}
\definecolor{promptborder}{RGB}{180,195,220}
\definecolor{promptheader}{RGB}{55,90,155}
\definecolor{modalbg}{RGB}{240,248,240}
\definecolor{modalborder}{RGB}{100,160,100}
\definecolor{stepbg}{RGB}{255,250,240}
\definecolor{stepborder}{RGB}{200,160,80}

\newtcolorbox{promptbox}[1]{
  breakable,
  colback=promptbg,
  colframe=promptborder,
  fonttitle=\bfseries\small,
  title=#1,
  coltitle=white,
  colbacktitle=promptheader,
  boxrule=0.6pt,
  arc=4pt,
  left=6pt, right=6pt, top=4pt, bottom=4pt,
  titlerule=0pt,
  toptitle=3pt, bottomtitle=3pt
}

\newtcolorbox{modalbox}[1]{
  breakable,
  colback=modalbg,
  colframe=modalborder,
  fonttitle=\bfseries\small,
  title=#1,
  coltitle=white,
  colbacktitle=modalborder!80!black,
  boxrule=0.6pt,
  arc=4pt,
  left=6pt, right=6pt, top=4pt, bottom=4pt,
  titlerule=0pt,
  toptitle=3pt, bottomtitle=3pt
}

\newtcolorbox{stepbox}[1]{
  breakable,
  colback=stepbg,
  colframe=stepborder,
  fonttitle=\bfseries\small,
  title=#1,
  coltitle=white,
  colbacktitle=stepborder!90!black,
  boxrule=0.6pt,
  arc=4pt,
  left=6pt, right=6pt, top=4pt, bottom=4pt,
  titlerule=0pt,
  toptitle=3pt, bottomtitle=3pt
}

\def\BibTeX{{\rm B\kern-.05em{\sc i\kern-.025em b}\kern-.08em
    T\kern-.1667em\lower.7ex\hbox{E}\kern-.125emX}}
\usepackage{balance}
\begin{document}
\title{Drone-PUR: From Visual Clue Discovery to Understanding and Reasoning in UAV Scenes}
\author{Jiahe Wu, Bing Cao, \emph{Member, IEEE}, Xingyue Guo, Ruipu Zhao, Beiming Yu, Jiekang Feng,\\
Qinghua Hu, \emph{Senior Member, IEEE}, Pengfei Zhu

\thanks{
This work was supported in part by Science and Technology Innovation Program of Xiongan New Area (Grant No.2025XAGG0039), in part by the National Natural Science Foundation of China (NSFC) under Grant 62436002, and in part by Tianjin Natural Science Funds for Distinguished Young Scholar under Grant 23JCJQJC00270.
\emph{(Corresponding author: Pengfei Zhu.)}

Jiahe Wu, Bing Cao, Ruipu Zhao, Jiekang Feng and Qinghua Hu are with the School of Artificial Intelligence, Tianjin University, Tianjin 300072, China.
(e-mail: wujiahe@tju.edu.cn; caobing@tju.edu.cn; ruipuzhao@tju.edu.cn, fengjk@tju.edu.cn, huqinghua@tju.edu.cn)

Xingyue Guo is with the College of Computer Science, VCIP, Nankai University, Tianjin 300350, China. (e-mail: guoxingyue@dbis.nankai.edu.cn)

Beiming Yu is with the School of Computer Science and Technology, Tianjin University, Tianjin 300072, China (e-mail: beimingyu@tju.edu.cn).


Pengfei Zhu is with the School of Automation, Southeast University, Nanjing 210096, China, also with the School of Artificial Intelligence, Tianjin University, Tianjin 300072, China, the Low-Altitude Intelligence Laboratory, Xiong’an National Innovation Center, Baoding, 071701, China, and Xiong’an Guochuang Lantian Technology Company Ltd., 
Baoding, 071701, China (e-mail: zhupengfei@tju.edu.cn).
}
}



\maketitle

\begin{abstract}
The reasoning capabilities of vision–language models (VLMs) can be limited by incomplete or inaccurate visual observations in the general domain. In unmanned aerial vehicle (UAV) scenes, small targets, subtle attributes, and changing viewpoints further complicate perception and understanding. Therefore, we introduce Drone-PUR, a dataset and benchmark organized around visual clue discovery, interpretation, and integration, 
corresponding to the three capability dimensions of perception, understanding, and reasoning.
Reasoning samples are designed to require models to combine multiple visual clues to derive visually grounded conclusions. Its 11 tasks provide 175k training and 20k test samples spanning single images, image sequences, cross-modal images, and multiple viewpoints. 
All test samples undergo human review.
Evaluating 22 general-purpose VLMs reveals persistent limitations in spatial judgment and cross-modal and cross-view reasoning. 
Across models, stronger perception is associated with better reasoning, while single-task fine-tuning further reveals selective bidirectional transfer between these capabilities. To examine domain adaptation, we develop UAV-PUR from Qwen3-VL through staged fine-tuning with auxiliary supervision. Its 4B and 8B variants improve overall multiple-choice accuracy by 18.33 and 17.52 percentage points over their pretrained counterparts, respectively. The dataset and project resources are available at \url{https://drone-pur.github.io/}.


\end{abstract}


\section{Introduction}

\IEEEPARstart{R}{ecent} advances in multimodal learning have enabled vision--language models (VLMs) to combine visual perception with language understanding, supporting complex reasoning and task execution. These capabilities enable unmanned aerial vehicle (UAV) applications, including navigation~\cite{TravelUAV}, anomaly understanding~\cite{A2Seek}, and object search~\cite{ji2026towards}.

However, VLMs' strong reasoning capabilities do not guarantee reliable inference from images.
Previous studies~\cite{eagle,VisuRiddles,MathVerse} on general-domain visual reasoning have shown that fine-grained perception errors can undermine subsequent inference.
Compared with general-domain imagery, UAV images often contain small, densely distributed targets that occupy few pixels, yet their subtle attributes may be critical for identifying objects and interpreting their states and behaviors.
Changes in UAV viewpoints can also alter object appearance and spatial reference frames, leading to spatial disorientation. 
These characteristics can obscure key visual clues and hinder their semantic and spatial alignment, leaving subsequent reasoning grounded in incomplete or incorrect visual information. 
Cross-view, cross-modal, or cross-frame inputs further require models to associate complementary clues across observations. 
Evaluating reasoning in UAV scenes therefore requires examining the visual capabilities that underpin it, including how models discover and interpret visual clues and integrate observations to support their inferences.

Recent UAV benchmarks have expanded evaluation from perception to spatial understanding, collaborative understanding, and task planning. UAVBench~\cite{UAVIT-1M} and UAVQA-Bench~\cite{UAVQA-Bench} assess image-level and region-level understanding, while MM-UAVBench~\cite{Mm-uavbench} extends capability evaluation to cognition and planning. Other studies investigate spatial and spatiotemporal understanding~\cite{SpatialSky,SpatialUAV,open3d,urbanvideo}, as well as embodied tasks~\cite{CityEQA,Embodiedcity}. Collaborative understanding is further investigated through multi-drone observations and cross-platform image pairs~\cite{Aircopbench,LinkS2,Urbench}. 
Despite this expanding scope of evaluation, further systematic study is needed to clarify how VLMs differ in discovering, interpreting, and integrating UAV visual clues, how these abilities relate, and whether domain-specific training can improve them jointly and enable cross-task transfer.

\begin{figure*}[tb]
\centering
\includegraphics[width=0.95\linewidth]{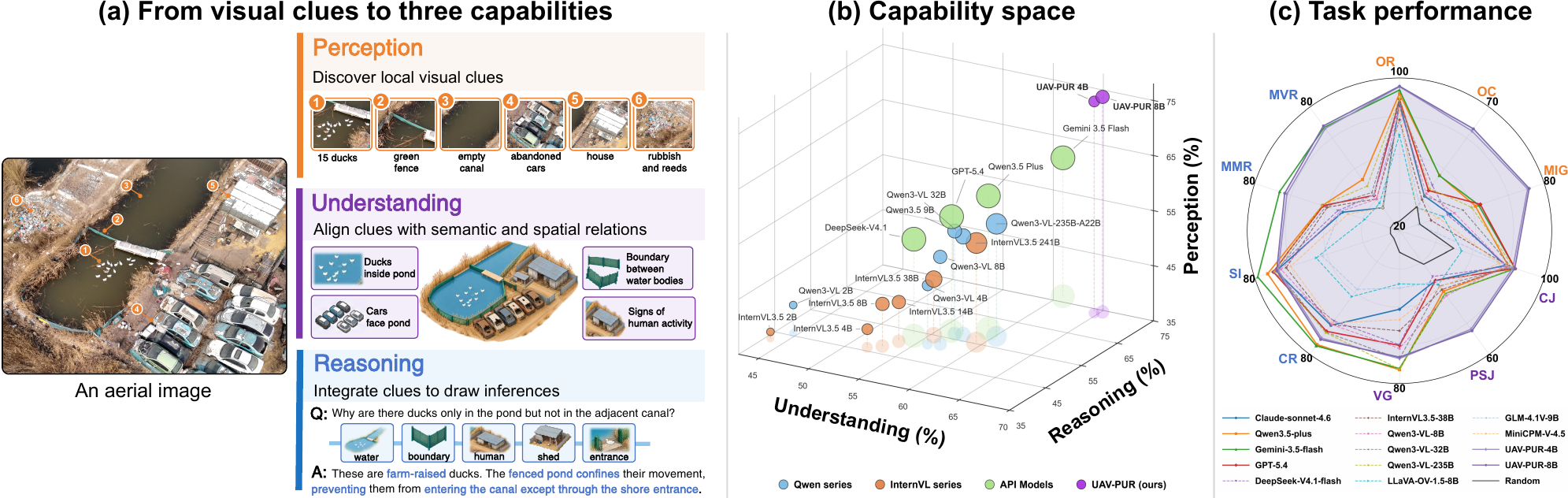}
\caption{The capability framework and model evaluation on Drone-PUR.  (a) The three capability dimensions illustrated through a shared scene. 
(b) Capability scores across three dimensions; marker sizes schematically indicate total parameter counts. 
(c) Task-level performance across the ten multiple-choice tasks.}

\label{fig_1}
\end{figure*}

To investigate these questions, we introduce Drone-PUR, a UAV dataset and benchmark organized across three capability dimensions: \textbf{P}erception, \textbf{U}nderstanding, and \textbf{R}easoning. 
These dimensions are distinguished by the roles that visual clues play in forming an answer. 
In this work, we define a visual clue as a task-relevant image region containing an individual object or a group of objects and associated with a specific modality, viewpoint, and capture time. 
Perception discovers and recognizes visual clues and extracts basic information, including categories, attributes, counts, and locations. 
Understanding aligns these clues with the semantic conditions and spatial concepts. 
Reasoning connects multiple visual clues through logical inference to form visually grounded explanations or conclusions. Fig.~\ref{fig_1}(a) illustrates these dimensions within the same UAV scene: models identify ducks and fences, interpret the enclosure structure, spatial layout, and signs of human activity, and combine these observations to explain why all the ducks are concentrated in the pond. This organization allows us to examine both dimension-specific challenges and interactions among capabilities.

Drone-PUR instantiates the three capability dimensions through 11 tasks spanning single images, image sequences, paired RGB and thermal infrared (TIR) observations, and synchronized UAV views. 
It integrates images from eight UAV datasets we previously released with 66,940 newly collected images, yielding 175k training samples and 20k test samples across diverse scenes. 
Over 90\% of the annotated task-relevant regions occupy less than 1\% of the image area, highlighting the fine-grained visual demands of Drone-PUR. 
The four reasoning tasks in Drone-PUR cover causal explanations and situational inference within single images, as well as clue integration across modalities and viewpoints. 
Each pure reasoning sample requires connecting multiple relevant clues to support the conclusion.
Sample construction combines VLM-based generation, cross-model verification, and human review, involving over 800 person-hours of manual effort. 
The Position Spatial Judgment (PSJ) task is fully human-annotated, and all test samples undergo human review. 
The benchmark supports multiple-choice evaluation on ten tasks and open-ended detection, grounding, and counting, assessing both answer selection and direct quantitative prediction.

We further develop UAV-PUR, a unified adaptation baseline initialized from pretrained Qwen3-VL models~\cite{qwen3} at the 4B and 8B scales. 
Its three-stage training comprises perception adaptation, joint capability learning, and final consolidation with language-generation supervision alone. 
During the first two stages, we introduce three auxiliary heads to provide additional geometric and numerical supervision, helping align visual representations with UAV task requirements. 
At inference, all tasks use only the original language modeling head. 

We evaluate 22 general-purpose VLMs on Drone-PUR. Within the same model family, larger models generally achieve stronger performance across the three capability dimensions (Fig.~\ref{fig_1}(b)). However, most models still struggle with PSJ and multi-view reasoning (MVR) (Fig.~\ref{fig_1}(c)). Spatial judgment does not improve consistently with model scale, and models exhibit systematic directional preferences in their predictions. 
Analyses of relationships across tasks and dimensions show a strong correlation between perception and reasoning across models ($r=0.95$). Single-task fine-tuning further reveals selective bidirectional transfer: learning to discover and interpret clues can improve reasoning, 
while training on clue integration tasks can, in turn, enhance perception.
Clue masking and modality/view ablations highlight the importance of essential local clues and complementary observations for reasoning.

UAV-PUR improves performance across all three dimensions, particularly in target matching and multi-source grounding. 
Its 4B and 8B variants raise overall multiple-choice accuracy by 18.33 and 17.52 percentage points over their pretrained counterparts, respectively. 
UAV-PUR still trails the best-performing API model on pure reasoning tasks, while fine-grained spatial interpretation, reliable open-ended prediction, and cross-source clue integration remain challenging.

Our main contributions are summarized as follows:
\begin{itemize}
\item We introduce Drone-PUR, a UAV dataset and benchmark organized around a perception--understanding--reasoning framework defined by the roles of visual clues. 
Its training and evaluation resources for 11 tasks support systematic assessment and domain adaptation.

\item We investigate relationships among UAV visual capabilities through model evaluation, single-task fine-tuning, and ablations. These analyses reveal strong perception--reasoning associations, selective bidirectional capability transfer, and the importance of local clues in reasoning.

\item We assess the benefits of UAV-domain adaptation using UAV-PUR, a baseline combining staged fine-tuning with auxiliary supervision. Joint training across tasks improves all three capability dimensions while revealing persistent limitations in reasoning and spatial understanding.

\end{itemize}

\section{Related Work}

\subsection{Vision--Language Tasks and Benchmarks in UAV Scenes}

Compared with general imagery, UAV scenes often contain smaller targets, less discernible attributes, and more variable viewing geometry, motivating dedicated datasets for aerial perception.
Earlier UAV benchmarks~\cite{visdrone,uavdt,HIT-UAV} evaluate vision models' ability to detect small objects under varying conditions.
UAVScenes~\cite{uavscenes} provides image and LiDAR annotations for UAV semantic segmentation.
Language-conditioned benchmarks further examine whether models can associate visual details with fine-grained descriptions and contextual relations. 
RefDrone~\cite{sun2026refdrone} and SkyFind~\cite{wang2026skyfind} study referring expression grounding~\cite{PSRN, CPGR} in UAV scenes, where small objects, scale variations, and clutter complicate target identification.
DVGBench~\cite{DVGBench}
introduces grounding queries that require scene-specific knowledge and inference. 
UAV-SVG~\cite{UAV-SVG} extends visual grounding to targets across aerial video frames. 
These tasks emphasize the alignment of linguistic descriptions with local visual details, contextual relationships, and target identity. 
Drone-PUR organizes four grounding settings and evaluates single-image, temporal, cross-view, and joint RGB--TIR grounding within a unified capability framework.

VLMs broaden UAV evaluation from visual perception to scene understanding, spatial intelligence, and planning.
MM-UAVBench~\cite{Mm-uavbench} assesses perception, cognition, and planning in UAV scenarios, while UAVBench~\cite{UAVIT-1M} covers image-level and region-level vision--language tasks. 
UAVQA-Bench~\cite{UAVQA-Bench} evaluates UAV image understanding through multiple-choice and visual grounding tasks. 
UrbanVideo-Bench~\cite{urbanvideo} assesses spatiotemporal understanding in videos. 
Other benchmarks investigate spatial understanding and three-dimensional spatial relations~\cite{SpatialSky,open3d,SpatialUAV}, or mathematical reasoning~\cite{AVI-Math}. SpatialSky~\cite{SpatialSky} evaluates environmental perception and scene understanding using data from UAVScenes~\cite{uavscenes}.
Collaborative perception and cross-view alignment are also examined through multi-drone observations and paired UAV--satellite inputs~\cite{Aircopbench,LinkS2,Urbench}. 
Alongside these efforts to evaluate UAV capabilities, Drone-PUR explicitly defines reasoning as connecting visual clues through logical inference to support an answer. 
Its training and evaluation resources support analyses of adaptation gains, cross-task performance transfer, and remaining challenges after UAV-domain fine-tuning.

\subsection{Evaluation of Visual Perception and Reasoning}

Evaluating VLMs involves assessing both visual perception and higher-level reasoning, motivating benchmarks that characterize individual capabilities and reveal limitations obscured by aggregate question-answering scores~\cite{mme, mmbench, vqa_survey}. 
BLINK~\cite{blink} evaluates perception-demanding tasks, including visual correspondence, relative depth estimation, and multi-view judgments. 
VSI-Bench~\cite{VSI} examines spatial understanding and memory from sequential video observations, while MMSI-Bench~\cite{MMSI} evaluates multi-image spatial reasoning involving correspondence, spatial transformations, and spatial logic. 
MLLM-CompBench~\cite{Mllm-compbench} studies comparative reasoning across image pairs, covering attributes, states, quantities, and spatial relations.
The role of visual perception in reasoning is also evident across domains. 
In mathematical and geometric problem solving, EAGLE~\cite{eagle} and MathVerse~\cite{MathVerse} show that inaccurate perception of diagram elements propagates into erroneous reasoning paths.
VisuRiddles~\cite{VisuRiddles} identifies fine-grained perception, rather than reasoning capacity, as the primary bottleneck in abstract visual reasoning.

Beyond individual task scores, compositional questions and hierarchical capability frameworks help examine how basic visual observations support semantic interpretation and more complex reasoning. 
GQA~\cite{GQA} uses scene graphs to construct questions that compose object attributes and relations. 
LENS~\cite{LENS} examines relationships among multimodal capabilities at different levels using shared general-domain images, reducing variation in visual content across tasks. 
CHOICE~\cite{Choice} develops a hierarchical evaluation of perception and reasoning in remote sensing (RS) at different granularities. 
HopChain~\cite{Hopchain} synthesizes logically dependent multi-hop questions in which earlier observations constrain the entities or relations examined in subsequent steps, requiring repeated grounding in visual clues. 
These studies motivate examining how visual information is organized and utilized throughout inference. 
Compared with general visual scenes, UAV imagery presents denser and smaller targets, subtler attributes, and greater viewpoint variation, complicating clue acquisition and association.
We organize UAV visual capabilities around three dimensions: perception discovers clues, understanding aligns them with semantic and spatial meaning, and reasoning combines them through logical inference. 

\subsection{Vision--Language Models and Domain Adaptation}

VLMs~\cite{qwen3,internvl35, qwen35} integrate visual representations with the language understanding and reasoning capabilities of large language models, enabling diverse visual tasks to be specified and solved through natural-language instructions. 
However, differences in object scale, viewing geometry, and visual context can limit the performance of general-purpose VLMs in UAV and other scenes.
Domain-specific instruction tuning addresses these differences by adapting pretrained VLMs to the visual distributions and task requirements of the target domain. GeoChat~\cite{Geochat}, the LHRS-Bot~\cite{Lhrs-bot, Lhrs-bot-nova}, and SkyEyeGPT~\cite{Skyeyegpt} construct RS instruction data to align visual representations with language and support RS-specific tasks. 
In the UAV domain, 
UAVIT-1M~\cite{UAVIT-1M} provides large-scale instruction data for vision--language alignment.
UAV-VL-R1~\cite{Uav-vl-r1} and Sky-VLM~\cite{SpatialSky} are trained on UAV-domain data using supervised fine-tuning and reinforcement learning to enhance scene understanding. 
These studies demonstrate the value of domain-specific data and training for aerial imagery. 
Following this direction, we provide UAV-PUR as a unified fine-tuning baseline on Drone-PUR to quantify the benefits of UAV-domain adaptation and examine challenges in reasoning.

\section{Drone-PUR Dataset} 
\begin{figure*}[tb]
\centering
\includegraphics[width=1\linewidth]{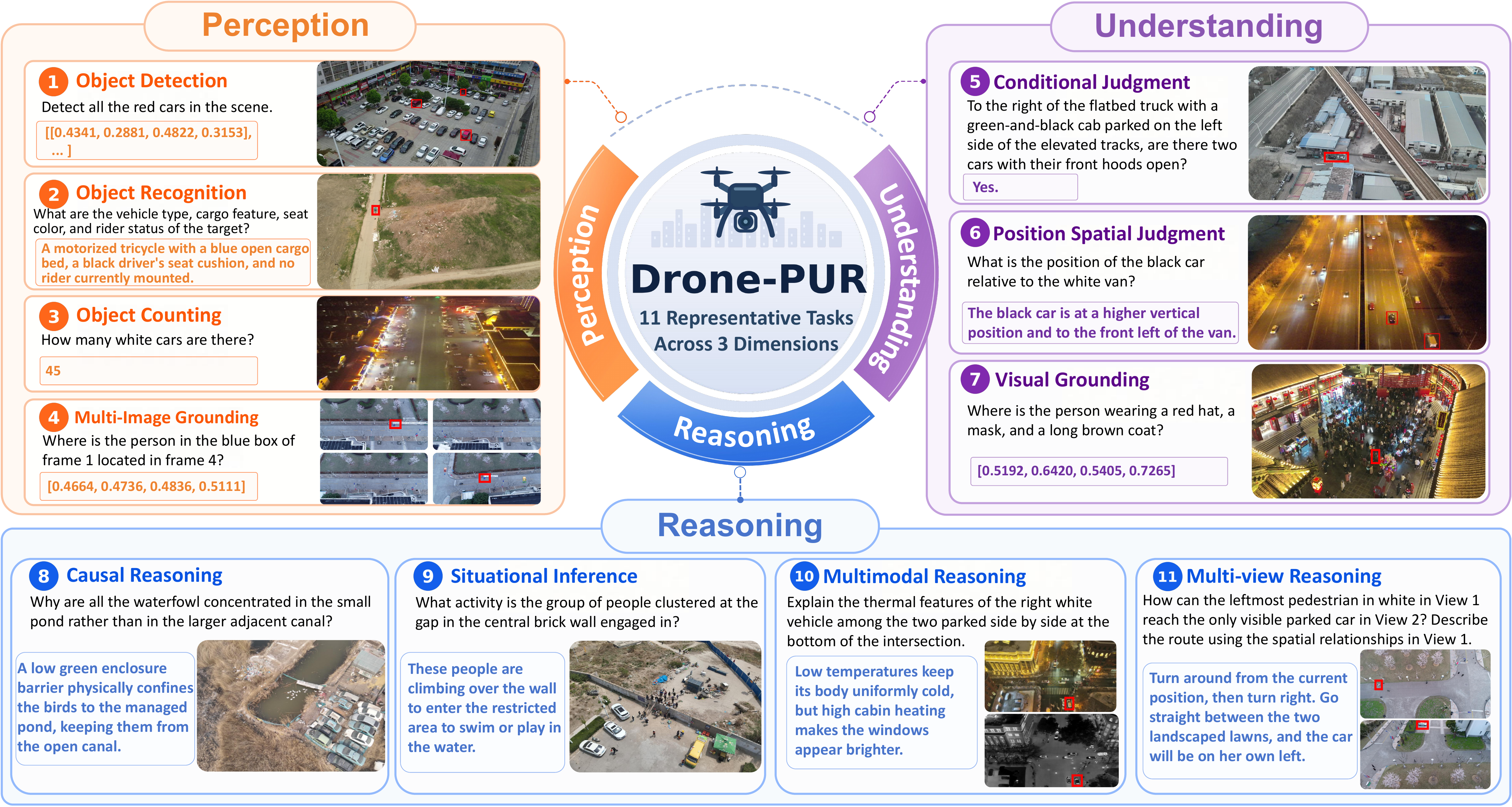}
\caption{Task overview of the Drone-PUR dataset across perception, understanding, and reasoning. Representative questions and reference answers are shown.
} 
\label{fig_2}
\end{figure*}

Drone-PUR is designed to evaluate how vision–language models perceive, understand, and reason about visual clues in UAV imagery. It comprises 11 tasks organized into three capability dimensions, with 175,213 training samples and 20,051 test samples. Fig.~\ref{fig_2} provides an overview of the task taxonomy and examples. 
The dataset covers single images, image sequences, paired views, and RGB–TIR inputs, supporting systematic evaluation of diverse visual capabilities and model adaptation to the UAV domain.

\subsection{Task Taxonomy and Design Principles} 

We define a visual clue as a task-relevant image region containing one or more objects and associated with a specific modality, viewpoint, and capture time. 
Task demands range from discovering visual clues and interpreting their semantic and spatial meaning to connecting them through logical inference.
Drone-PUR organizes these requirements into three capability dimensions: perception, understanding, and reasoning. 
These dimensions characterize the primary demands of each task while allowing the underlying capabilities to overlap.

\subsubsection{\textbf{Perception}}
Perception focuses on discovering objects and extracting their features and comprises the following tasks:

\noindent\textbf{Object Detection (OD).}
Given a textual condition, predict the bounding boxes of all objects satisfying the specified category or attribute requirements.

\noindent\textbf{Object Recognition (OR).}
Given a designated object, identify its category and combined visual attributes.

\noindent\textbf{Object Counting (OC).}
Predict the number of objects belonging to a specified category in the image.

\noindent\textbf{Multi-Image Grounding (MIG).}
Given a template image with a target bounding box and three query images from the same sequence, locate the target in a specified query image.

OD and OC require searches across the scene to identify all qualifying instances. 
OR and MIG focus on designated targets, extracting distinctive features for basic category and attribute recognition or target identity matching. Together, these tasks assess the acquisition of object clues that support subsequent interpretation and inference.

\subsubsection{\textbf{Understanding}}
Understanding evaluates whether models can align visual clues with the semantic conditions and spatial concepts specified by a question. Its tasks include:

\noindent\textbf{Conditional Judgment (CJ).}
Determine whether an image satisfies a compound textual condition involving object attributes and relationships.

\noindent\textbf{Visual Grounding (VG).} 
Localize the unique object that matches a textual description. Descriptions may combine visual attributes, relative spatial relations, and ordinal positions.

\noindent\textbf{Position Spatial Judgment (PSJ).} 
It comprises five subtasks: vehicle heading (VH), pedestrian heading (PH), turning direction (TD), horizontal relative position (RP), and height comparison (HC).
VH and PH describe heading orientation in the image coordinate system.
TD and RP are defined with respect to the target's and reference object's own orientation, respectively, while 
HC compares the vertical levels occupied by objects, rather than their physical heights.

Building on perceptual capabilities, CJ aligns visual content with compound textual conditions involving multiple objects. VG aligns specific object instances with discriminative descriptions, while PSJ interprets spatial concepts.

\subsubsection{\textbf{Reasoning}} 

Reasoning evaluates whether models can integrate visual clues and contextual relationships through logical inference to derive visually supported reasoning conclusions or answers to multi-source joint tasks.

\noindent\textbf{Causal Reasoning (CR).} 
Identify the explanation best supported by visual clues for an observed phenomenon.

\noindent\textbf{Situational Inference (SI).} 
Infer an activity, state, or situation by connecting visual clues and their contextual relationships.

\noindent\textbf{Multimodal Reasoning (MMR).} 
Integrate cross-modal clues from temporally aligned RGB-TIR images of the same scene.

\noindent\textbf{Multi-View Reasoning (MVR).} 
Establish spatial correspondences and connect complementary clues across two images captured simultaneously from different viewpoints.

\begin{figure*}[tb]
\centering
\includegraphics[width=1\linewidth]{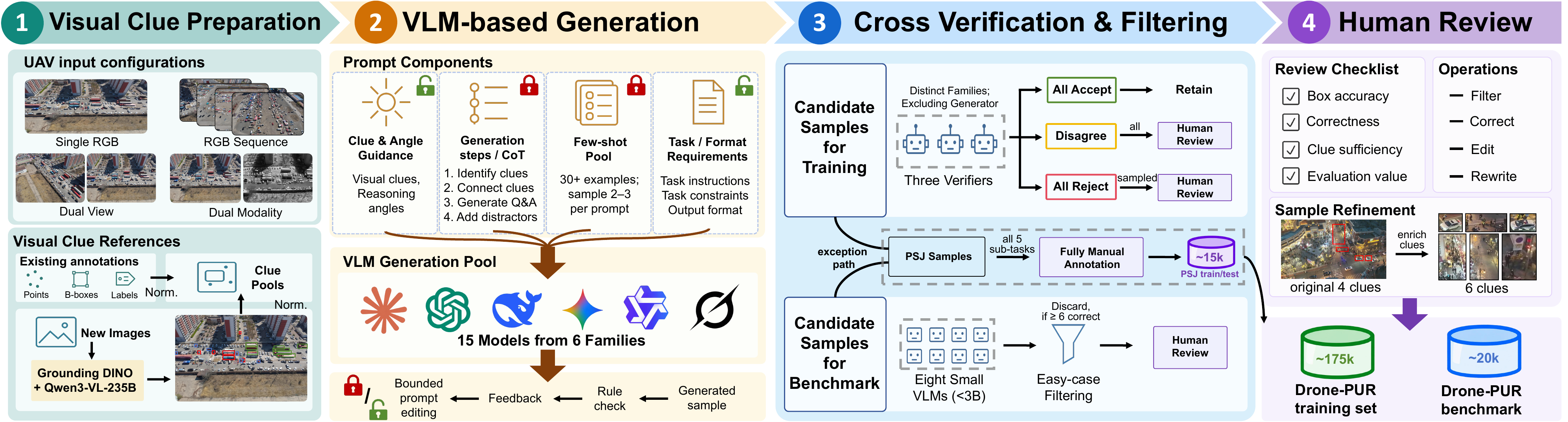}
\caption{Construction pipeline of Drone-PUR, comprising clue preparation, VLM-based generation, cross-model verification and filtering, and human review.} 
\label{fig_3}
\end{figure*}

MMR and MVR distinguish different components of clue integration. MMR includes joint grounding (MMR-G), joint counting (MMR-C), and pure reasoning (MMR-R), while MVR includes cross-view grounding (MVR-G) and pure reasoning (MVR-R). 
MMR-G identifies and localizes a target using attributes from both modalities, MMR-C counts objects using clues from both modalities, and MVR-G locates the counterpart of a designated target in another view. 
Although their outputs are perceptual, these subtasks require integrating complementary clues across modalities or utilizing visual context to establish cross-view correspondences. They assess the joint perception and target correspondence needed for multi-source reasoning.
Each reasoning sample in CR, SI, MMR-R, and MVR-R requires integrating multiple visual clues, typically four to seven, to form a coherent explanation or inference.
The intended inference depends on relationships among these clues, which jointly constrain possible interpretations and support the answer.

\subsection{Data Sources and Annotations}

\begin{figure*}[tb]
\centering
\includegraphics[width=1\linewidth]{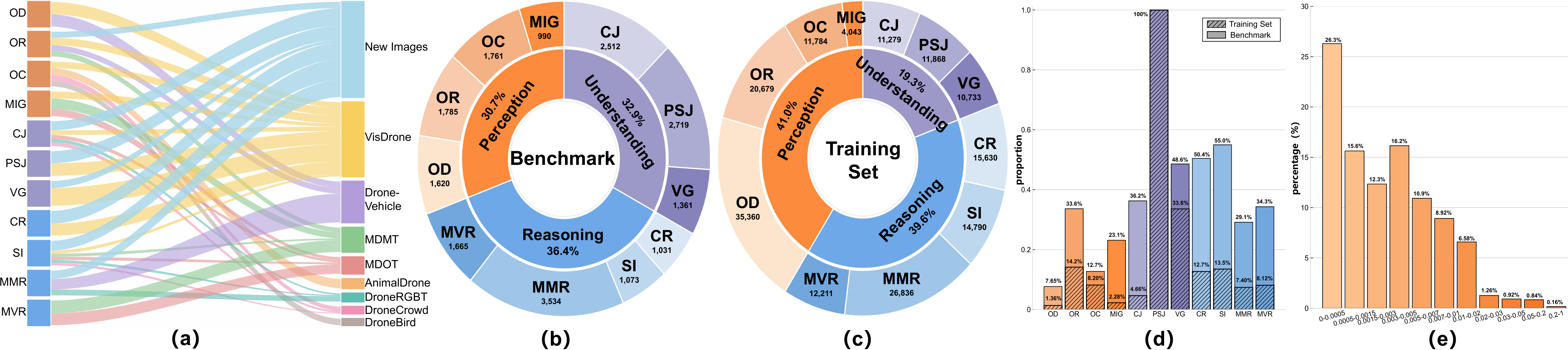}
\caption{Statistical overview of the Drone-PUR dataset. (a) Image sources by task. (b–c) Task distributions in the benchmark and training sets, respectively. (d) Proportions of human-authored or revised samples by task and split. (e) The distribution of region areas normalized by image area.} 
\label{fig_4}
\end{figure*}

Drone-PUR integrates images and annotations from eight UAV datasets previously released by our team: VisDrone~\cite{visdrone}, DroneVehicle~\cite{dronevehicle},  MDMT~\cite{MDMT}, MDOT~\cite{MDOT}, DroneRGBT~\cite{dronergbt}, DroneCrowd~\cite{dronecrowd}, AnimalDrone~\cite{animaldrone}, and DroneBird~\cite{dronebird}. 
Originally developed for scene-wide object detection and counting, as well as tracking, these datasets provide UAV imagery with bounding-box, point, and category annotations, but lack semantic descriptions.
We additionally introduce 66,940 newly collected images, extending coverage to urban, rural, industrial, forest, maritime, and anomalous scenes. 
Fig.~\ref{fig_4}(a) summarizes the image sources used by each task, with the newly collected images labeled as ``New Images''.
The newly collected data were captured at heights of 8--400~m. 
Following A2Seek~\cite{A2Seek}, we mask identifiable faces and license plates in newly collected low-altitude images unless consent has been obtained.
RGB resolutions range from $1920 \times 1080$ to $8000 \times 6000$,
while TIR resolutions range from $640 \times 512$ to $1280 \times 1024$.
The RGB--TIR pairs are temporally synchronized but exhibit spatial offsets.
Consequently, multimodal tasks require models to establish correspondences
from scene context while combining appearance and thermal information.
We repartition images from all sources into training and test sets, applying the same partition across all tasks. 
No images or frames from the same source video appear in both sets, 
avoiding frame- and video-level leakage.

We convert all bounding boxes in the dataset to a unified $[x_1, y_1, x_2, y_2]$ format, with coordinates normalized to $[0,1]$.
These annotations serve as target references, grounding answers, candidate options, as well as candidate visual clues for generating reasoning samples.
Selected partially occluded or truncated targets are retained to represent challenging UAV observations.
For newly collected images without existing annotations, we combine Qwen3-VL-235B~\cite{qwen3} and Grounding DINO~\cite{Groundingdino} to generate initial bounding-box annotations, extending category coverage to buildings, vegetation, facilities, and additional animal and vehicle categories. 

\subsection{Sample Construction and Quality Control}
Fig.~\ref{fig_3} illustrates the four-stage construction pipeline: clue preparation, VLM-based generation, cross-model verification and filtering, and human review. Training and test candidates follow different validation paths. 
Except for PSJ, training samples undergo model verification, with uncertain cases referred for human review, whereas all test samples undergo small-model filtering followed by comprehensive human review. 

\textbf{1) Clue preparation.} 
We organize images into task-specific input configurations and associate them with available object annotations. 
For understanding and reasoning samples, relevant bounding boxes provide explicit references to candidate visual clues.
For reasoning tasks, annotators further identify relevant clues and suggest possible directions of inference to guide visually grounded question generation.

\textbf{2) VLM-based generation.} 
We employ fifteen models from six model families: Gemini, Grok, Qwen, Claude, GPT, and DeepSeek. Each generation prompt contains four components: task requirements, step-by-step generation instructions, guidance on visual clues and reasoning angles, and few-shot examples. 
Examples are randomly sampled from a manually written pool to encourage diversity in content and phrasing.

For pure reasoning samples, step-by-step prompts guide the generator to first inspect the images, identify relevant visual clues, and select clues that jointly support an explanation or inference about the same phenomenon.
It then constructs the question, reference answer, and plausible distractors.
For reasoning candidates, the generation prompt requires the VLM to combine at least four visual clues to jointly support a conclusion. This requirement guides automated generation toward questions involving multiple observations. 
Images that lack meaningful reasoning potential are skipped. 
The generator also proposes a clue list, a reasoning chain, and explanations for the distractors as references for subsequent review.

Format and task-specific rule checks identify invalid candidates, such as those involving prohibited attributes, rule violations, or leading questions.
Based on this feedback, the model may revise the task-specific requirements and visual clue guidance in the prompt, with each update modifying fewer than 5\% of the prompt tokens. Core task constraints, generation steps, and the content of few-shot examples remain unchanged throughout this process.

\textbf{3) Cross-model verification and filtering.} 
Automatically generated training candidates are verified by three VLMs from distinct families in the generator pool, excluding the generator of each candidate.
The verifiers assess whether the sample is correct and unambiguous.
For pure reasoning candidates, they also assess whether the proposed clues are needed to support the intended answer. 
The clue lists and reasoning chains serve as review aids rather than validated reasoning annotations; acceptance is based on the question–answer pair and its visual support.
All training candidates unanimously accepted by the three verifiers are retained.
Candidates with conflicting judgments are submitted for human review. 
We also sample a small subset of unanimously rejected candidates for human review and possible revision to recover valid questions that the verifiers may have misjudged. The remaining unanimously rejected candidates are discarded.

Test candidates are filtered using eight small VLMs with fewer than 3B parameters. A candidate is discarded as an easy case when more than five VLMs answer it correctly. All remaining test candidates undergo human review.

\textbf{4) Human review.} 
Human reviewers assess samples across tasks for bounding-box accuracy, answer correctness, clue sufficiency, ambiguity, and evaluation value. 
They may accept, filter, edit, or rewrite samples. 
For selected reasoning samples, reviewers refine samples to construct more challenging ones by increasing the number of clues that must be considered jointly. 
Samples submitted for human review are independently assessed by at least two reviewers. Disputed samples are referred to a review panel and retained only when all panel members agree that the revised reference answer is visually supported.
All samples in the test set undergo human review. The cumulative annotation and review effort exceeds 800 person-hours.

Unlike the other ten tasks, all samples of PSJ are constructed entirely by human annotators.
We adopt manual annotation because prompted VLMs do not consistently generate valid spatial judgment samples. Its training and test sets contain 14,587 samples in total, covering all five subtasks.

\subsection{Dataset Statistics} 
Following the four-stage construction pipeline, Drone-PUR contains 168k images and 620k bounding-box annotations, including 200k boxes used in tasks other than OD. 
The training and test sets contain 175,213 and 20,051 samples, respectively, with all 11 tasks represented in both sets. 
Fig.~\ref{fig_4}(b) and (c) show the task distributions. 
Perception, understanding, and reasoning account for 30.7\%, 32.9\%, and 36.4\% of the benchmark, and 41.0\%, 19.3\%, and 39.6\% of the training set.
Fig.~\ref{fig_4}(d) reports the proportion of samples in the final dataset manually authored or substantially revised by human annotators. 
This includes manual question construction, correction, editing, and rewriting, but excludes VLM-generated samples accepted without modification.
In four pure reasoning tasks, human-authored or revised samples account for more than 50\% of each test set,
introducing variation in question content and reasoning difficulty.

To characterize the relative sizes of task-relevant visual regions, we deduplicate the bounding-box annotations used across tasks and normalize their areas by the corresponding image areas. 
The analysis includes boxes used for target references, grounding options, and visual clue guidance during reasoning sample generation. 
As shown in Fig.~\ref{fig_4}(e), 90.2\% of the boxes occupy less than 1\% of the image area. Regions occupying more than 5\% of the image area typically correspond to buildings and facilities. 
The predominance of small regions highlights the fine-grained visual demands of Drone-PUR: models are required to capture subtle local details to identify, interpret, and integrate visual clues.




\section{A Baseline for UAV-Domain Adaptation}

\begin{figure}[tb]
\centering
\includegraphics[width=1\linewidth]{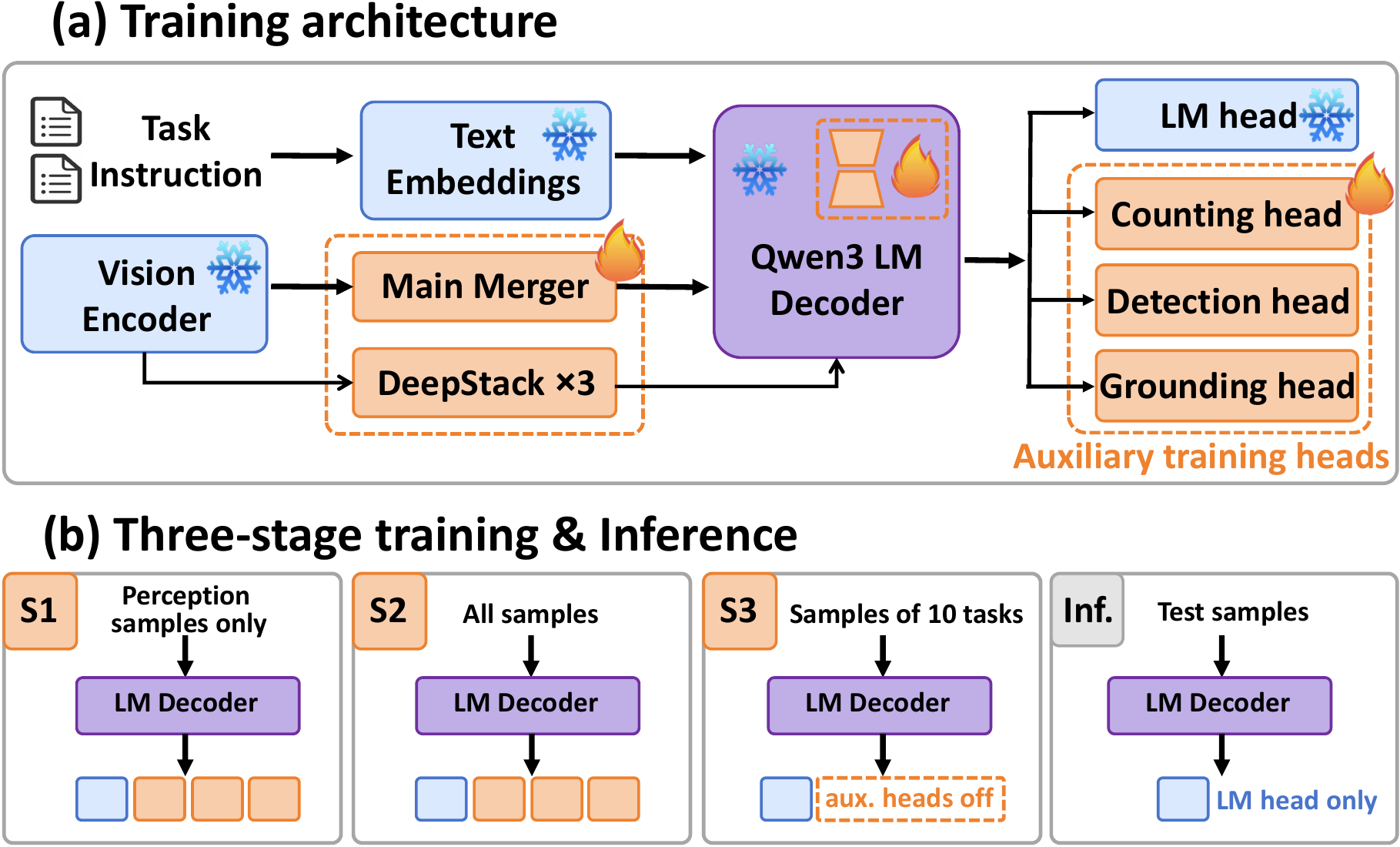}
\caption{Overview of UAV-PUR, illustrating the model and training architecture, three-stage fine-tuning with auxiliary supervision, and unified inference through the original language modeling head. Auxiliary heads are used only in the first two training stages.}
\label{fig_5}
\end{figure}

Building on Drone-PUR, we develop UAV-PUR, a unified domain adaptation baseline covering 11 UAV tasks with both multiple-choice answers and open-ended predictions. 
Built on standard components, UAV-PUR combines stage-wise fine-tuning with auxiliary supervision to provide a reference for assessing the benefits and limitations of joint training on Drone-PUR.
As illustrated in Fig.~\ref{fig_5}, the model receives text supervision throughout training, with auxiliary heads providing explicit geometric and numerical supervision for eligible open-ended samples in the first two stages.

We initialize UAV-PUR from pretrained Qwen3-VL~\cite{qwen3} models at the 4B and 8B scales.
Each input consists of one or more images, a task instruction, and a question. 
Multiple-choice samples additionally include candidate options.
We apply LoRA to the linear projection layers in the attention and feed-forward modules of the language model and fully fine-tune the main visual merger and three DeepStack mergers.
All other pretrained parameters remain frozen, including the visual encoder, text embeddings, and original language modeling head.
As in Qwen3-VL, the three DeepStack mergers project intermediate ViT features into the language model's hidden space for injection into its early layers. 
Updating the visual mergers adapts the mapping from frozen visual features to language-model representations for UAV scenes.

To provide additional supervision, we introduce three auxiliary heads: a grounding head, a counting head, and a detection head. 
For sample $i$, $H_i^{\mathrm{prompt}}$ denotes the final-layer hidden states of the multimodal prompt, and $h_i^{\mathrm{cond}}$ denotes its last non-padding token state. 
The grounding and counting heads each process $h_i^{\mathrm{cond}}$ using LayerNorm followed by a two-layer MLP with 512 hidden units and GELU activation. 
The grounding head converts sigmoid-activated center and scale parameters into valid normalized boxes within $[0,1]$, while the counting head uses Softplus to produce nonnegative count estimates.
The detection head adopts a two-layer DETR-style~\cite{DETR} decoder with 128 learnable object queries. 
Projected prompt states $H_i^{\mathrm{prompt}}$ serve as memory, and a projection of $h_i^{\mathrm{cond}}$ is added to each query. 
Separate branches predict bounding boxes 
and objectness scores for targets.
All auxiliary heads use only prompt representations, which cannot attend to subsequent answer tokens under the language model's causal mask. Auxiliary gradients flow through these shared representations to update the trainable parameters.

All samples receive autoregressive text supervision through the original language modeling head.
For input $x_i$ and target answer $y_i$ containing $T_i$ tokens,
the text loss is:
\begin{equation}
\label{eq:baseline_ce}
\mathcal{L}_{\mathrm{text},i}
=
-\frac{1}{T_i}
\sum_{t=1}^{T_i}
\log p_{\theta}
\left(y_{i,t}\mid x_i,y_{i,<t}\right).
\end{equation}

Token-level supervision $\mathcal{L}_{\mathrm{text},i}$ does not explicitly account for coordinate distances, box overlap, or numerical errors.
We therefore introduce auxiliary losses for eligible open-ended samples.
The grounding and counting heads predict a normalized bounding box $\hat{\mathbf{b}}_i$ and a nonnegative count $\hat{c}_i$ from $h_i^{\mathrm{cond}}$, respectively, with objectives:

\begin{equation}
\label{eq:baseline_box}
\mathcal{L}_{\mathrm{grd},i}
=
\frac{1}{4}
\left\|\hat{\mathbf{b}}_i-\mathbf{b}_i\right\|_1
+
1-\operatorname{GIoU}
\left(\hat{\mathbf{b}}_i,\mathbf{b}_i\right),
\end{equation}
\begin{equation}
\label{eq:baseline_count}
\mathcal{L}_{\mathrm{cnt},i}
=
\left|\hat{c}_i-c_i\right|,
\end{equation}
where $\mathbf{b}_i$ and $c_i$ are the GT bounding box and count, respectively.
Some grounding samples without a valid target use 
\texttt{not exist} as the correct answer and receive text supervision only.

For OD, Hungarian matching assigns predictions to ground-truth objects one-to-one using a cost combining L1 distance, GIoU, and objectness. The loss is defined as:
\begin{equation}
\label{eq:baseline_detection}
\mathcal{L}_{\mathrm{det},i}
=
5\mathcal{L}_{\mathrm{L1},i}^{\mathrm{match}}
+
2\mathcal{L}_{\mathrm{GIoU},i}^{\mathrm{match}}
+
\mathcal{L}_{\mathrm{obj},i},
\end{equation}
where the box terms are coordinate-summed L1 and $1-\mathrm{GIoU}$ losses, averaged over matched objects within each sample. 
The objectness loss uses binary cross-entropy with weights of 1 for matched queries and 0.1 for unmatched queries.

The overall objective at stage $s$ is defined as:
\begin{equation}
\label{eq:baseline_total}
\mathcal{L}^{(s)}
=
\frac{1}{B}
\sum_{i=1}^{B}
\left[
\mathcal{L}_{\mathrm{text},i}
+
\sum_{k\in\{\mathrm{grd},\mathrm{cnt},\mathrm{det}\}}
m_{i,k}^{(s)}\lambda_k\mathcal{L}_{k,i}
\right],
\end{equation}
where $B$ is the batch size, $\lambda_k$ controls the weight
of auxiliary loss $k$, and $m_{i,k}^{(s)}$ indicates whether that loss is enabled for sample $i$ at stage $s$. Each sample activates at most one auxiliary loss.
Multiple-choice samples receive text supervision only.
Training proceeds through three stages, with each stage initialized from the preceding stage's checkpoint.

\noindent\textbf{Stage 1: Perception adaptation.}
We first train on OD, OC, MIG, and OR to adapt the model to UAV imagery and strengthen its perceptual ability to discover visual clues.
OC and MIG include both multiple-choice and open-ended formats.
The text loss is supplemented with the applicable grounding, counting, or detection loss.

\noindent\textbf{Stage 2: Joint capability learning.}
To jointly learn perception, understanding, and reasoning, we train all 
eleven tasks 
together with six open-ended variants: VG, MIG, MMR-G, and MVR-G for grounding, and OC and MMR-C for counting.
Auxiliary losses remain active for eligible open-ended samples.

\noindent\textbf{Stage 3: Consolidation with text supervision.}
We remove all auxiliary heads and continue training on a mixture of the ten multiple-choice tasks and the six open-ended variants.
Only the text loss is retained, so $m_{i,k}^{(3)}=0$ for every sample and auxiliary objective.
The language LoRA parameters and visual mergers remain trainable, aligning optimization with generative inference while retaining open-ended samples to help preserve grounding and counting capabilities.

At inference, all auxiliary heads are discarded.
UAV-PUR only uses the original language modeling head to generate option labels or open-ended predictions according to the input instruction.
This preserves a unified interface across tasks without additional task-specific branches.

\section{Experiments}
\subsection{Experimental Setup}
\noindent\textbf{Evaluation protocol.}
We conduct evaluation on Drone-PUR, the benchmark constructed in this work, using two evaluation formats: multiple-choice
answering and open-ended prediction.
The multiple-choice evaluation covers the ten tasks, excluding OD, with a total of 18,431 test samples.
Evaluated models receive the unified task instructions, required images, question, and candidate options, and are evaluated by accuracy.
CJ provides two options per question, the remaining perception and understanding tasks provide four, and the reasoning tasks provide five. 
Each question has one designated reference option. For explanatory and situational questions, this option represents the interpretation judged to be best supported by the visible evidence among the provided alternatives.

Open-ended evaluation removes candidate options and requires models to directly generate coordinates or numerical values.
For VG, MIG, MMR-G, and MVR-G, models predict normalized bounding boxes.
We report grounding accuracy at an intersection-over-union (IoU) threshold of 0.5 (Acc@0.5), considering a prediction correct when its IoU with the reference box exceeds 0.5.
For samples without a valid target in the image, correctness is determined by whether the model predicts \texttt{not exist}.
For OC and MMR-C, models directly predict object counts.
We report mean absolute error (MAE) and mean relative accuracy (MRA) following VSI-Bench~\cite{VSI}. MAE measures the average absolute error, while MRA averages accuracy across multiple relative-error tolerances:
\begin{equation}
\label{eq6}
\mathrm{MRA}
=
\frac{1}{N|\mathcal{C}|}
\sum_{i=1}^{N}
\sum_{\theta\in\mathcal{C}}
\mathbf{1}
\left[
\frac{|\hat{c}_i-c_i|}{c_i}<1-\theta
\right],
\end{equation}
where $\hat{c}_i$ and $c_i$ denote the predicted and ground-truth counts, respectively, and $N$ is the number of evaluated samples. We use $\mathcal{C}=\{0.5,0.55,\ldots,0.95\}$, corresponding to relative-error tolerances from 5\% to 50\% in 5\% increments.

For OD, we report micro-averaged F1 at an IoU threshold of 0.5, denoted as: 
\begin{equation}
\label{eq7}
\mathrm{Micro\text{-}F1@0.5}
=\frac{2\sum TP}{2\sum TP+\sum FP+\sum FN},
\end{equation}
where predicted and ground-truth boxes are matched one-to-one within each query using IoU $>0.5$, maximizing the number of valid matches. True positives, false positives, and false negatives are then aggregated over the entire OD test set.

\noindent\textbf{Implementation details.}
We fine-tune the pretrained Qwen3-VL-4B~\cite{qwen3} and Qwen3-VL-8B to obtain UAV-PUR-4B and UAV-PUR-8B, respectively.
Each of the three stages is trained for one epoch using AdamW with a weight decay of 0.01, cosine learning-rate decay, and a warm-up ratio of 0.05.
The LoRA rank, $\alpha$, and dropout are set to 32, 64, and 0.05, with a global batch size of 32.
The peak learning rates for the language LoRA parameters are $2\times10^{-5}$, $2\times10^{-5}$, and $1\times10^{-5}$ in Stages 1, 2, and 3, respectively.
The corresponding peak learning rates for the visual mergers are $2\times10^{-6}$, $2\times10^{-6}$, and $1\times10^{-6}$,
while active auxiliary heads use a peak learning rate of $1\times10^{-4}$.
Tasks are sampled in proportion to their training-set sizes, with approximate multiple-choice-to-open-ended sampling ratios of $1:1$, $3:1$, and $5:1$ across the three stages, respectively.
The grounding, counting, and detection loss weights are 0.1, 0.01, and 0.1.
Auxiliary losses are enabled according to the training stage and sample type.

\subsection{Benchmark Results and Analysis}

\begin{table*}[!t]
  \centering
  \caption{ Performance on the Drone-PUR multiple-choice benchmark. 
  We report API-evaluated models, locally evaluated models, and fine-tuned models. 
  The best and second-best scores within each model category are shown in bold and underlined, respectively. 
  }
  {
  \renewcommand{\arraystretch}{0.99}
  \resizebox{1\linewidth}{!}{
  \begin{tabular}{lccccccccccc}
    \toprule
    \multirow{2}{*}{Method}
& \multicolumn{3}{c}{\cellcolor{orange!27} Perception}
& \multicolumn{3}{c}{\cellcolor{blue!25} Understanding}
& \multicolumn{4}{c}{\cellcolor{blue!35!cyan!30} Reasoning}
& \multirow{2}{*}{Overall $\uparrow$} \\
~ & OR & OC & MIG & CJ & PSJ & VG & CR & SI & MMR & MVR & \\
\midrule
Random
& 25.00 & 25.00 & 25.00
& 50.00 & 25.00 & 25.00
& 20.00 & 20.00 & 20.00 & 20.00 & 26.43 \\

Text-only
& 35.07 & 22.37 & 25.86
& 54.62 & 28.14 & 27.04
& 28.90 & 26.75 & 23.43 & 21.86 & 30.16 \\

Human
& 98.21 & - & 99.70
& 92.28 & 95.59 & 98.46
& 96.80 & 97.39 & 92.44 & 98.92 & - \\

\hline
\multicolumn{12}{c}{\cellcolor{gray!20}\small\textit{API-Evaluated Models}}\\

Claude-sonnet-4.6~\cite{claude_sonnet46}
& 86.61 & 29.64 & 38.48
& 78.54 & 31.34 & 48.86
& 64.99 & 70.46 & 41.17 & 27.93 & 50.38 \\

Qwen3.5-plus~\cite{qwen35}
& \underline{91.76} & \textbf{39.07} & 49.49
& 80.25 & \underline{34.94} & \textbf{74.21}
& \underline{74.98} & \underline{74.28} & \underline{50.06} & \underline{42.46} & \underline{58.80} \\

Gemini-3.5-flash~\cite{gemini35flash}
& \textbf{93.67} & \underline{39.01} & \underline{50.91}
& \textbf{83.92} & \textbf{36.34} & \underline{73.70}
& \textbf{75.56} & \textbf{78.66} & \textbf{68.96} & \textbf{69.61} & \textbf{66.09} \\

GPT-5.4~\cite{gpt54}
& 88.91 & 32.20 & \textbf{51.72}
& \underline{83.84} & 31.30 & 64.07
& 68.09 & 71.76 & 49.80 & 34.29 & 55.87 \\

DeepSeek-V4.1-flash~\cite{xu2026deepseek}
& 86.27 & 34.58 & 37.37
& 79.78 & 33.10 & 63.78
& 62.75 & 71.67 & 43.24 & 34.47 & 53.22 \\

Grok-4.3~\cite{grok43}
& 79.22 & 24.87 & 26.77
& 76.04 & 28.69 & 35.42
& 51.79 & 57.97 & 37.58 & 36.76 & 45.49 \\

\hline
\multicolumn{12}{c}{\cellcolor{gray!20}\small\textit{Locally Evaluated Models}}\\
Qwen3-VL-2B~\cite{qwen3}
& 65.21 & 26.69 & 32.53
& 69.82 & 30.23 & 36.08
& 47.33 & 60.02 & 36.08 & 32.13 & 43.22 \\

InternVL3.5-2B~\cite{internvl35}
& 56.36 & 23.23 & 32.22
& 67.95 & 28.80 & 37.25
& 45.97 & 46.13 & 33.22 & 33.33 & 40.31 \\

MiniCPM-V-4.0-4B~\cite{minicpmv40_modelcard}
& 65.55 & 26.75 & 30.71
& 71.22 & 28.72 & 45.41
& 53.93 & 53.59 & 34.24 & 28.83 & 43.16 \\

Qwen3-VL-4B~\cite{qwen3}
& 77.98 & 26.12 & 35.86
& 81.25 & 34.17 & 62.09
& 65.47 & 67.29 & 41.48 & 33.45 & 51.23 \\

InternVL3.5-4B~\cite{internvl35}
& 63.53 & 25.10 & 30.10
& 76.91 & 33.32 & 51.36
& 55.67 & 52.10 & 37.66 & 30.51 & 45.48 \\

MiMo-VL-7B~\cite{yue2025mimo}
& 73.89 & 24.36 & 33.74
& 79.74 & 32.18 & 41.29
& 52.76 & 62.16 & 35.29 & 29.31 & 45.94 \\

LLaVA-OV-1.5-8B~\cite{llava-ov}
& 70.25 & 24.65 & 35.35
& 54.34 & 30.89 & 38.28
& 50.15 & 52.75 & 31.35 & 30.21 & 40.46 \\

Qwen3-VL-8B~\cite{qwen3}
& 84.09 & 30.83 & 40.20
& 78.70 & \textbf{36.41} & 65.32
& 64.50 & 68.31 & 42.05 & 32.37 & 52.74 \\

InternVL3.5-8B~\cite{internvl35}
& 74.01 & 23.96 & 33.23
& 76.87 & 32.51 & 55.03
& 57.61 & 51.63 & 40.24 & 30.63 & 47.29 \\

MiniCPM-V-4.5-8B~\cite{Minicpm-v4.5}
& 83.75 & 24.65 & 33.33
& 75.20 & 33.32 & 53.27
& 63.72 & 64.03 & 38.71 & 27.99 & 48.60 \\

GLM-4.1V-9B~\cite{glm-4.1}
& 77.09 & 26.35 & 24.34
& 81.13 & 31.67 & 40.04
& 46.36 & 50.14 & 42.08 & 34.59 & 46.68 \\

Qwen3.5-9B~\cite{qwen35}
& \textbf{90.03} & \underline{31.69} & 40.91
& 81.25 & 32.59 & 69.07
& 65.76 & 66.82 & 48.08 & \textbf{39.40} & 55.27 \\

InternVL3.5-14B~\cite{internvl35}
& 72.49 & 23.96 & 33.54
& 79.38 & 31.96 & 57.97
& 59.75 & 55.73 & 43.01 & 32.73 & 48.72 \\

Qwen3-VL-32B~\cite{qwen3}
& 88.07 & 29.70 & \underline{44.04}
& \underline{81.89} & 33.50 & \underline{69.58}
& \underline{68.57} & \underline{70.18} & \underline{48.64} & 35.74 & \underline{55.45} \\

InternVL3.5-38B~\cite{internvl35}
& 78.15 & 28.11 & 30.00
& 81.25 & 34.90 & 57.90
& 64.69 & 66.08 & 47.88 & 32.43 & 51.94 \\

Qwen3-VL-235B~\cite{qwen3}
& \underline{89.36} & \textbf{32.99} & \textbf{47.88}
& \textbf{83.48} & \underline{36.34} & \textbf{73.99}
& \textbf{68.77} & \textbf{71.20} & \textbf{50.17} & \underline{39.10} & \textbf{57.72} \\

\hline
\multicolumn{12}{c}{\cellcolor{gray!20}\small\textit{Models Fine-tuned on Drone-PUR}}\\


Qwen-\textit{sft}-4B
& 94.17 & 51.68 & 68.99
& 82.84 & 48.80 & 65.61
& 71.58 & 69.25 & 63.92 & 67.51 & 67.49 \\


\textbf{UAV-PUR-4B}
& \underline{95.46} & \underline{59.23} & \underline{72.63}
& 82.68 & \underline{50.28} & \textbf{69.51}
& \underline{71.68} & \underline{70.18} & 65.73 & \underline{69.01} & \underline{69.56} \\


Qwen-\textit{sft}-8B
& 95.29 & 57.30 & 69.70
& \underline{83.56} & 49.98 & 64.00
& 71.19 & 69.62 & \underline{65.87} & 67.15 & 68.67 \\



\textbf{UAV-PUR-8B}
& \textbf{95.69} & \textbf{60.19} & \textbf{73.33}
& \textbf{83.68} & \textbf{50.86} & \underline{68.92}
& \textbf{72.36} & \textbf{70.64} & \textbf{66.72} & \textbf{70.33} & \textbf{70.26} \\

\bottomrule
  \end{tabular}
  }
  }
\label{tab_1}
\end{table*}

\noindent\textbf{Evaluated models and reference settings.}
We evaluate 22 general-purpose VLMs: six through APIs and sixteen locally.
The API-evaluated models are Claude-sonnet-4.6~\cite{claude_sonnet46}, Qwen3.5-plus~\cite{qwen35}, Gemini-3.5-flash~\cite{gemini35flash}, GPT-5.4~\cite{gpt54}, DeepSeek-V4.1-flash~\cite{xu2026deepseek} and Grok-4.3~\cite{grok43}. The locally evaluated models span the Qwen, InternVL, MiniCPM, MiMo, LLaVA, and GLM families and cover a range of model sizes. 
We also report the performance of Qwen-\textit{sft} and UAV-PUR at both the 4B and 8B scales. 
Qwen-\textit{sft} is jointly fine-tuned on Drone-PUR for three epochs using only cross-entropy loss over answer tokens. It uses the same overall sampling ratio of multiple-choice to open-ended samples and the same trainable backbone modules as UAV-PUR.
We also include three reference baselines for multiple-choice evaluation: Random denotes the expected accuracy under uniform random selection; Text-only retains the questions, task instructions, and answer options but removes all images, reporting the average accuracy of three API models~\cite{qwen35,gemini35flash,gpt54} under this setting; Human reports the average accuracy of four evaluators who use image zoom tools when answering.

\noindent\textbf{Overall performance.}
Table~\ref{tab_1} summarizes performance across the ten multiple-choice tasks. 
PSJ, MMR and MVR accuracies are aggregated across their subtasks by sample count, and Overall is the sample-weighted accuracy across all ten tasks.
Among general-purpose VLMs, Gemini-3.5-flash achieves the highest overall accuracy of 66.09\%, exceeding the strongest locally evaluated model, Qwen3-VL-235B, by 8.37 percentage points. 
Performance varies substantially across tasks: the best general-purpose accuracy reaches 93.67\% on OR but only 39.07\% on OC and 36.41\% on PSJ, with the latter falling 59.18 points below human performance.
On MVR, Gemini-3.5-flash reaches 69.61\%, while all other general-purpose models score below 50\%. Only 7.8\% of the test samples of MVR were generated by Gemini-family models without human revision; the rest were human-edited or generated by other VLMs and approved by human reviewers. 
Text-only answering achieves 30.16\% overall, only 3.73 points above random guessing and substantially below performance with image inputs. 
This suggests that questions and answer options alone provide limited cues for answering, highlighting the importance of visual clues.
Within the Qwen3-VL and InternVL series, overall accuracy increases with model size, although task-level gains vary. Some reasoning tasks benefit consistently from scaling: SI accuracy increases across the Qwen3-VL series, while CR accuracy increases across the InternVL3.5 series. In contrast, PSJ performance shows no consistent improvement with scale: Qwen3-VL-8B achieves 36.41\%, compared with 33.50\% for 32B and 36.34\% for 235B.
We additionally evaluate two 7B-scaled UAV-adapted models, LLaVA1.5-UAV~\cite{UAVIT-1M} and Sky-VLM~\cite{SpatialSky}, which achieve overall accuracies of 42.79\% and 45.20\%, respectively, 
potentially reflecting differences in their training task formulations.

\begin{figure*}[!t]
\centering
\includegraphics[width=0.89\linewidth]{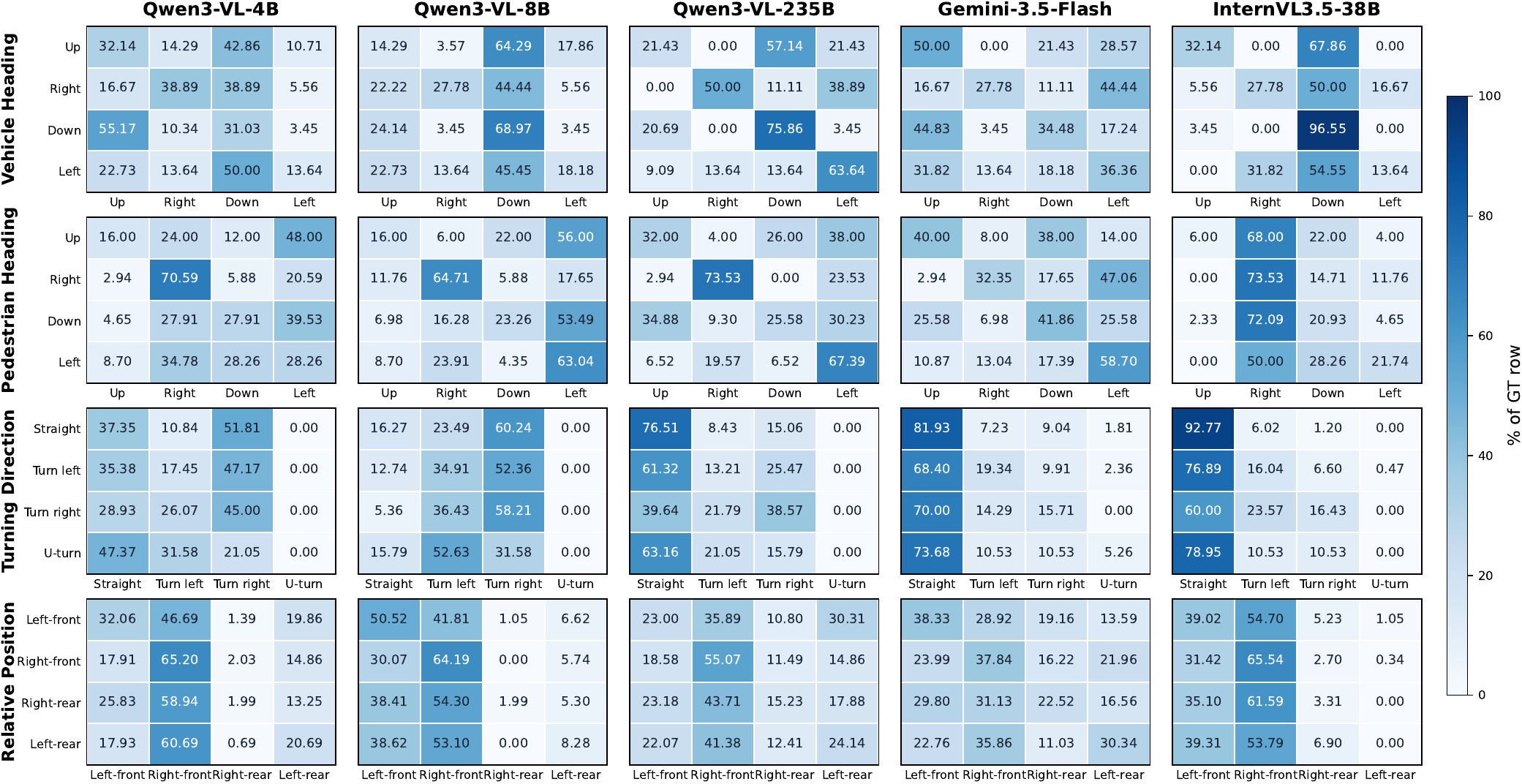}
\caption{Confusion matrices of five representative VLMs on four PSJ subtask groups. Results are shown for cardinal vehicle headings, cardinal pedestrian headings, turning directions, and four-quadrant relative positions. Each row is normalized by the number of samples in its ground-truth direction.}
\label{fig_6}
\end{figure*}

\begin{table}[!t]
\centering
\caption{Performance comparison on the subtasks of MMR and MVR.}
\label{tab_2}
\begingroup
\renewcommand{\arraystretch}{0.9}
\begin{tabular}{@{}
  >{\raggedright\arraybackslash}p{0.26\linewidth}
  *{5}{@{}>{\centering\arraybackslash}p{0.14\linewidth}}
@{}}
\toprule
Method & MMR-G & MMR-C & MMR-R & MVR-G & MVR-R \\
\midrule
Claude-4.6       & 45.09 & 26.00 & 75.61 & 21.27 & 47.41 \\
Qwen3.5-plus     & 64.16 & 30.68 & 76.10 & 36.10 & 61.08 \\
Gemini-3.5       & 86.62 & 50.35 & 85.85 & 68.57 & 72.64 \\
GPT-5.4          & 63.42 & 29.74 & 78.70 & 27.80 & 53.30 \\
DeepSeek-V4.1    & 39.64 & 31.56 & 82.76 & 25.71 & 60.14 \\
Grok-4.3         & 45.00 & 20.96 & 69.11 & 30.06 & 56.37 \\
\midrule
Qwen3-VL-2B      & 32.29 & 28.45 & 64.72 & 28.53 & 42.69 \\
InternVL3.5-2B   & 28.08 & 28.16 & 57.40 & 29.57 & 44.34 \\
MiniCPM-V-4.0    & 29.40 & 26.00 & 66.67 & 23.77 & 43.63 \\
Qwen3-VL-4B      & 52.35 & 21.66 & 75.12 & 30.22 & 42.92 \\
InternVL3.5-4B   & 41.70 & 24.06 & 67.48 & 25.71 & 44.58 \\
MiMo-VL-7B       & 39.64 & 18.91 & 72.20 & 24.01 & 44.81 \\
LLaVA-OV-1.5-8B  & 32.54 & 18.44 & 64.88 & 25.79 & 43.16 \\
Qwen3-VL-8B      & 54.00 & 21.31 & 76.10 & 27.64 & 46.23 \\
InternVL3.5-8B   & 42.28 & 29.22 & 66.83 & 27.07 & 41.04 \\
MiniCPM-V-4.5    & 35.18 & 28.45 & 74.15 & 22.72 & 43.40 \\
GLM-4.1V-9B      & 55.33 & 21.60 & 72.85 & 31.27 & 44.34 \\
Qwen3.5-9B       & 62.59 & 28.34 & 74.31 & 35.78 & 50.00 \\
InternVL3.5-14B  & 47.15 & 31.15 & 67.80 & 29.01 & 43.63 \\
Qwen3-VL-32B     & 59.79 & 30.62 & 76.75 & 32.72 & 44.58 \\
InternVL3.5-38B  & 57.97 & 30.39 & 76.59 & 28.61 & 43.63 \\
Qwen3-VL-235B    & 67.13 & 26.00 & 83.90 & 34.17 & 53.54 \\
\midrule
\textbf{UAV-PUR-4B} & 67.30 & 60.42 & 77.40 & 72.44 & 58.96 \\
\textbf{UAV-PUR-8B} & 69.61 & 60.36 & 78.70 & 73.81 & 60.14 \\
\bottomrule
\end{tabular}
  \label{tab_2}
\endgroup
\end{table}

\begin{table}[!t]
\centering
\caption{Performance comparison on the five subtasks of PSJ.}
\label{tab_3}
\begingroup
\renewcommand{\arraystretch}{0.9}
\begin{tabular}{@{}
  >{\raggedright\arraybackslash}p{0.26\linewidth}
  *{6}{@{}>{\centering\arraybackslash}p{0.12\linewidth}}
@{}}
\toprule
Method & VH & PH & TD & RP & HC & AVG. \\
\midrule
Claude-4.6      & 23.40 & 33.14 & 35.89 & 30.26 & 33.73 & 31.34 \\
Qwen3.5-plus    & 33.43 & 30.79 & 33.38 & 36.57 & 41.42 & 34.94\\
Gemini-3.5      & 36.21 & 38.42 & 32.79 & 36.23 & 47.34 & 36.34 \\
GPT-5.4         & 30.08 & 29.91 & 38.40 & 26.51 & 41.42 & 31.30 \\
DeepSeek-V4.1   & 34.82 & 39.59 & 28.36 & 33.84 & 30.18 & 33.10\\
Grok-4.3        & 25.91 & 34.90 & 28.21 & 27.54 & 31.95 & 28.69 \\
\midrule
Qwen3-VL-2B     & 27.30 & 24.05 & 29.84 & 32.40 & 35.50 & 30.23 \\
InternVL3.5-2B  & 23.68 & 21.41 & 26.14 & 32.74 & 37.87 & 28.80 \\
MiniCPM-V-4.0   & 23.68 & 25.51 & 27.77 & 32.57 & 23.08 & 28.72 \\
Qwen3-VL-4B     & 32.87 & 29.03 & 33.23 & 35.72 & 40.24 & 34.17 \\
InternVL3.5-4B  & 31.48 & 25.81 & 39.14 & 31.46 & 42.01 & 33.32 \\
MiMo-VL-7B      & 28.13 & 28.45 & 29.84 & 35.38 & 35.50 & 32.18 \\
LLaVA-OV-1.5-8B & 27.58 & 27.27 & 27.18 & 36.32 & 22.49 & 30.89 \\
Qwen3-VL-8B     & 34.82 & 29.33 & 39.00 & 36.74 & 41.42 & 36.41 \\
InternVL3.5-8B  & 28.41 & 26.98 & 37.67 & 31.97 & 35.50 & 32.51 \\
MiniCPM-V-4.5   & 26.18 & 26.39 & 35.75 & 35.29 & 39.05 & 33.32 \\
GLM-4.1V-9B     & 26.74 & 24.63 & 31.02 & 34.78 & 37.28 & 31.67 \\
Qwen3.5-9B      & 36.77 & 29.33 & 26.00 & 35.81 & 34.32 & 32.59 \\
InternVL3.5-14B & 26.18 & 21.11 & 37.67 & 32.65 & 38.46 & 31.96 \\
Qwen3-VL-32B    & 39.00 & 27.86 & 37.81 & 29.92 & 40.83 & 33.50 \\
InternVL3.5-38B & 33.15 & 31.09 & 34.56 & 35.55 & 43.20 & 34.90 \\
Qwen3-VL-235B   & 35.65 & 39.30 & 38.85 & 32.74 & 46.75 & 36.34 \\
\midrule
\textbf{UAV-PUR-4B} & 63.79 & 71.26 & 48.15 & 39.73 & 60.95 & 50.28 \\
\textbf{UAV-PUR-8B} & 77.72 & 64.52 & 46.97 & 40.15 & 56.21 & 50.86 \\
\bottomrule
\end{tabular}
  \label{tab_3}
\endgroup
\end{table}

Fine-tuning on Drone-PUR improves performance across all ten tasks at both model scales. UAV-PUR-4B and UAV-PUR-8B achieve overall accuracies of 69.56\% and 70.26\%, improving over their pretrained Qwen3-VL counterparts by 18.33 and 17.52 percentage points, respectively. 
Large gains occur on perception and understanding tasks, with UAV-PUR-8B improving OC and MIG by 29.36 and 33.13 points, respectively. These improvements indicate that domain-specific training effectively narrows the performance gap associated with UAV imagery. 
However, spatial judgment remains challenging: UAV-PUR-8B achieves only 50.86\% on PSJ, falling 44.73 percentage points below human performance.
Improvements on CR and SI are more modest, and UAV-PUR-8B remains below Gemini-3.5-flash on both tasks; Qwen3-VL-235B also retains an advantage on SI. Thus, domain-specific fine-tuning does not uniformly close the gaps in reasoning performance, suggesting that integrating multiple visual clues may further benefit from the reasoning capacity afforded by larger models.

\noindent\textbf{MMR and MVR subtask performance.} 
Table~\ref{tab_2} reports multiple-choice performance on the grounding (G), counting (C), and reasoning (R) subtasks of MMR and MVR. 
Gemini-3.5-flash substantially outperforms other general-purpose models on several subtasks.
On MMR-G, it even surpasses the fine-tuned models, UAV-PUR. 
In contrast, all other general-purpose models score below 40\% on multimodal counting and cross-view grounding, highlighting persistent bottlenecks in joint perception under multi-source inputs.
Fine-tuning substantially improves these weaker components: UAV-PUR-8B achieves 60.36\% on MMR-C and 73.81\% on MVR-G. 
However, Gemini-3.5-flash leads on pure reasoning subtasks, reaching 85.85\% on MMR-R and 72.64\% on MVR-R, compared with 78.70\% and 60.14\% for UAV-PUR-8B. 
These results highlight uneven gains across components: substantial improvements in grounding and counting do not necessarily translate into comparable improvements in reasoning.


\noindent\textbf{PSJ subtask performance.} Table~\ref{tab_3} reports performance on the five PSJ subtasks. 
General-purpose VLMs achieve a sample-weighted accuracy of at most 36.41\%. Their performance is low across all five subtasks, with accuracies generally around 30\%; some models even perform near or below the 25\% random-guessing baseline on VH and PH.
These VLMs struggle to make reliable spatial judgments in UAV imagery.
Fine-tuning raises this score to 50.28\% for UAV-PUR-4B and 50.86\% for UAV-PUR-8B, but improvements vary substantially across spatial attributes. 
UAV-PUR-8B reaches 77.72\% on VH, whereas RP remains challenging: its accuracy increases only from 36.74\% to 40.15\%.
These results show that improvements in overall PSJ accuracy do not translate into uniform gains across spatial attributes.

\noindent\textbf{Directional preferences.}
To further examine the error patterns in PSJ, we analyze the directional confusion matrices of five representative general-purpose VLMs.
As shown in Fig.~\ref{fig_6}, 
the confusion matrices reveal systematic error patterns, with predictions tending toward particular directions.
In VH, upward headings are frequently mistaken for downward ones: Qwen3-VL-8B, Qwen3-VL-235B, and InternVL3.5-38B make this error in 64.29\%, 57.14\%, and 67.86\% of cases, respectively.
PH further reveals model-specific preferences: for upward headings, Qwen3-VL-4B and 8B predict left in 48.00\% and 56.00\% of cases, while InternVL3.5-38B predicts right in 68.00\%. 
Thus, the same ground-truth heading can elicit different directional preferences across models, including confusion between vertical and horizontal headings.
In TD, Qwen3-VL-4B and 8B frequently mistake left turns for right turns, while Qwen3-VL-235B, Gemini, and InternVL3.5-38B more often predict straight travel. 
For Gemini and InternVL3.5-38B, straight predictions account for at least 60\% of samples in every ground-truth category, indicating a strong preference for straight travel even when the actual movement involves a turn.
U-turn recognition is particularly weak: all three Qwen models and InternVL3.5-38B have zero recall, while Gemini achieves only 5.26\%.
RP shows a recurring tendency to mistake rear positions for front positions: Qwen3-VL-4B, 8B, 235B, and InternVL3.5-38B classify right-rear targets as right-front in 58.94\%, 54.30\%, 43.71\%, and 61.59\% of cases, respectively. 
Although Gemini exhibits less confusion along this particular direction, its right-rear recall remains only 22.52\%. 
Overall, these results indicate pronounced directional preferences in the models' spatial predictions.


\begin{figure}[!t]
\centering
\includegraphics[width=0.9\linewidth]{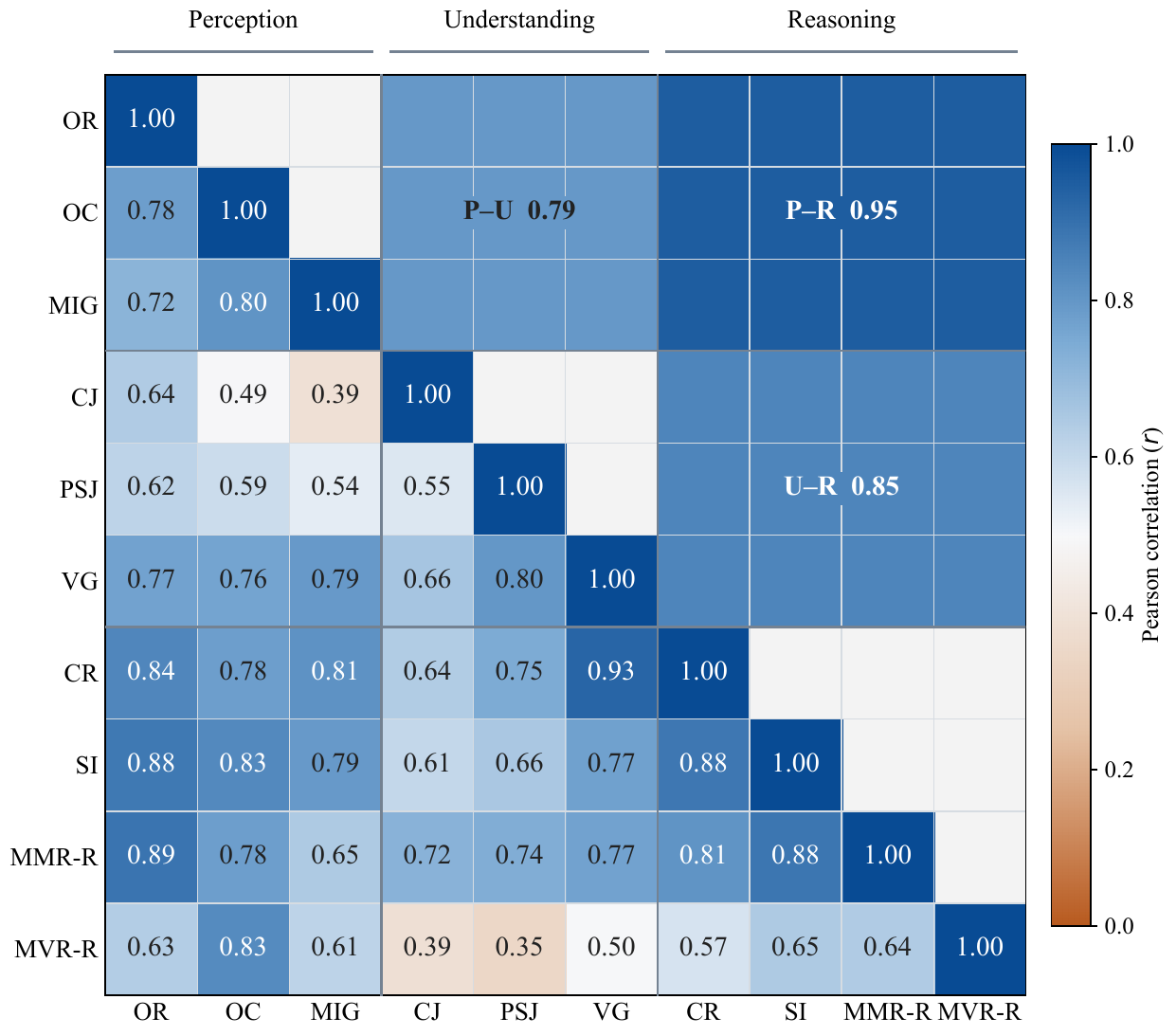}
\caption{Pearson correlations across tasks and capability dimensions.}
\label{fig_7}
\end{figure}

\noindent\textbf{Task and dimension correlations.}
We compute Pearson correlations across the 22 general-purpose VLMs in Table~\ref{tab_1}, excluding fine-tuned models.
For MMR and MVR, we use scores on their pure reasoning subsets, MMR-R and MVR-R, instead of aggregate task scores to exclude grounding and counting components.
Figure~\ref{fig_7} presents task-level correlations in the lower triangle and dimension-level correlations in the upper blocks, with each dimension score computed as the sample-weighted mean of its constituent task scores.
Perception and reasoning exhibit the strongest dimension-level correlation ($r=0.95$), followed by understanding and reasoning ($r=0.85$) and perception and understanding ($r=0.79$). 
Strong task-level associations also emerge between VG and CR ($r=0.93$) and between OR and MMR-R ($r=0.89$). These patterns suggest that models with stronger recognition and grounding performance also tend to perform better on reasoning tasks that require extracting and combining visual clues. However, the associations vary substantially across tasks. MVR-R correlates weakly with CJ, PSJ, and VG (lower than 0.51), but strongly with OC ($r=0.83$). 
Thus, performance on individual tasks provides differing indications of cross-view reasoning performance. This variation may reflect the additional demands of establishing correspondences and integrating clues across views.

\begin{figure}[tb]
\centering
\includegraphics[width=0.9\linewidth]{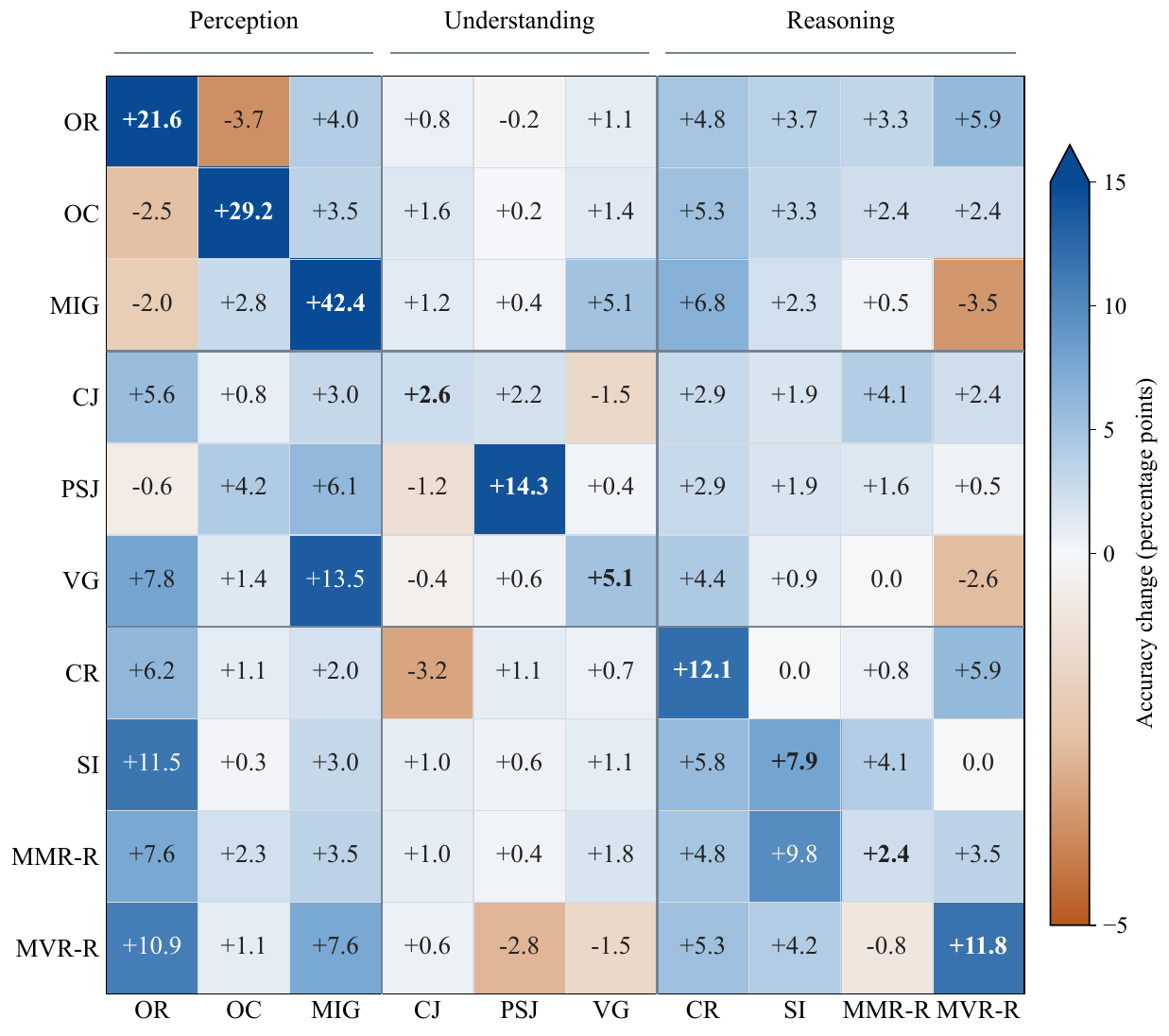}
\caption{Cross-task transfer following single-task fine-tuning on Drone-PUR.}
\label{fig_8}
\end{figure}

\noindent\textbf{Cross-task capability transfer.} 
While the correlations above describe performance associations across models, they do not establish whether training on one task benefits another. 
To examine such transfer, we independently fine-tune Qwen3-VL-4B for three epochs on the multiple-choice training samples of each source task and evaluate the resulting models across all ten target tasks. For MMR and MVR, both training and evaluation are restricted to their pure reasoning subtasks, MMR-R and MVR-R.
Figure~\ref{fig_8} reports accuracy changes relative to the original Qwen3-VL-4B on the same evaluation samples, with rows indicating training tasks and columns indicating evaluation tasks. Diagonal entries measure improvements on the training task itself, while off-diagonal entries measure transfer to other tasks. 
Perception training generally benefits reasoning, yielding positive changes in 11 of the 12 source--target pairs. Transfer also occurs in the reverse direction: training on each of the four reasoning tasks improves all three perception tasks, with SI and MVR-R training increasing OR accuracy by 11.5 and 10.9 percentage points, respectively. 
These bidirectional gains are consistent with the shared requirement to extract and use visual clues across the two dimensions. 

However, belonging to the same dimension does not guarantee positive transfer. Among perception tasks, OR and OC training improve MIG, but both OC and MIG training reduce OR accuracy. In contrast, transfer among reasoning tasks is predominantly positive, with MMR-R training improving SI by 9.8 percentage points, suggesting shared skills in combining visual clues. MVR-R exhibits more selective transfer. Although OR and CR training each improve MVR-R by 5.9 percentage points, MIG and VG training reduce its accuracy by 3.5 and 2.6 points. Conversely, MVR-R training improves perception but reduces PSJ and VG accuracy by 2.8 and 1.5 points. 
These asymmetric effects indicate that transfer involving cross-view reasoning is selective and does not consistently benefit other tasks.
Finally, the substantial within-task gains on perception tasks, together with their mixed cross-task transfer, indicate that strong task-specific adaptation does not necessarily translate into broad performance improvements.

\begin{table}[!t]
  \centering
      \caption{Open-Ended Detection and Grounding Performance.}
  {
  \renewcommand{\arraystretch}{1}

  \resizebox{1\linewidth}{!}{
  \begin{tabular}{lccccc}
    \toprule
    {Method}  & OD & MIG & VG & MMR-G & MVR-G\\
\hline
{Qwen3-VL-2B} & 15.33 & 0.40 & 14.40 & 3.14 & 0.32\\ 
{InternVL3.5-4B} & 2.61 & 0.00 & 7.27 & 0.83 & 10.96\\ 
{Qwen3-VL-4B}  & 29.50 & 4.04 & 43.86 & 23.20 & 9.43\\ 
{InternVL3.5-8B}  & 16.61 & 0.00 & 12.42 & 0.17 & 2.26\\ 
{Qwen3-VL-8B}  & 11.20 & 1.41 & 49.01 & 26.34 & 9.75\\ 
{InternVL3.5-14B}  & 11.42 & 0.00 & 14.70 & 2.15 & 12.81\\ 
{Qwen3-VL-32B}  & 51.02 & 7.37 & 16.53 & 32.20 &  16.68 \\ 
Qwen3-VL-235B  & 40.77 & 9.80 & 53.93 & 47.23 &  17.57 \\
Gemini-3.5-flash  & 21.35 & 34.34 & 27.99 & 56.98 & 44.56\\ 
\midrule
Qwen-\textit{sft}-4B & 36.90 & 6.57 & 42.76 & 33.20 & 26.11 \\ 
Qwen-\textit{sft}-8B & 18.73 & 17.37 & 48.42 & 31.38 & 42.22\\ 
UAV-PUR-4B   & 59.68 & 36.16 & 48.27 & 46.74 & 51.49 \\ 
UAV-PUR-8B   & 62.57 & 37.17 & 57.83 & 53.18 & 50.85\\ 

\bottomrule
  \end{tabular}}}
  \label{tab_4}
\end{table}

\begin{table}[!t]
  \centering
  \caption{Open-ended counting performance and underestimation rates.}
  \label{tab_5}
  \renewcommand{\arraystretch}{1}
  \resizebox{\linewidth}{!}{
  \begin{tabular}{lcc|cc|c}
    \toprule
    \multirow{2}{*}{Method}
      & \multicolumn{2}{c|}{OC}
      & \multicolumn{2}{c|}{MMR-C}
      & \multirow{2}{*}{\shortstack{Avg. \\Under(\%)}} \\
    & MAE$\downarrow$ & MRA(\%)$\uparrow$
    & MAE$\downarrow$ & MRA(\%)$\uparrow$ & \\
    \midrule
    Qwen3-VL-2B       & 97.43 & 10.90 & 15.35 & 36.23 & 87.07 \\
    Qwen3-VL-4B       & 76.41 & 12.08 &  9.29 & 48.53 & 84.76 \\
    Qwen3-VL-8B       & 69.20 & 18.64 &  9.02 & 51.04 & 80.11 \\
    InternVL3.5-8B    & 78.83 & 11.64 & 13.59 & 36.23 & 68.98 \\
    Qwen3.5-9B        & 73.58 & 26.52 &  8.96 & 57.70 & 82.69 \\
    LLaVA-OV-1.5      & 74.95 & 17.04 & 12.51 & 34.07 & 89.25 \\
    MiniCPM-V-4.5     & 63.11 & 29.01 & 10.44 & 47.56 & 69.92 \\
    InternVL3.5-14B   & 78.91 & 12.37 & 12.79 & 34.21 & 67.31 \\
    Qwen3-VL-32B      & 59.71 & 24.72 &  8.67 & 56.25 & 76.85 \\
    InternVL3.5-38B   & 86.17 & 11.06 & 12.89 & 38.28 & 68.95 \\
    Gemini-3.5-flash  & 72.65 & 23.22 &  6.03 & 63.14 & 87.54 \\
    \midrule
    Qwen-\textit{sft}-4B & 38.33 & 48.31 & 6.90 & 62.05 & 74.15 \\
    Qwen-\textit{sft}-8B & 58.59 & 54.05 & 5.83 & 65.88 & 47.38 \\
    UAV-PUR-4B          & 22.67 & 71.01 & 3.98 & 76.25 & 54.46 \\
    UAV-PUR-8B          & 23.75 & 70.03 & 4.22 & 74.28 & 58.45 \\
    \bottomrule
  \end{tabular}
  }
\end{table}

\begin{figure}[tb]
\centering
\includegraphics[width=0.99\linewidth]{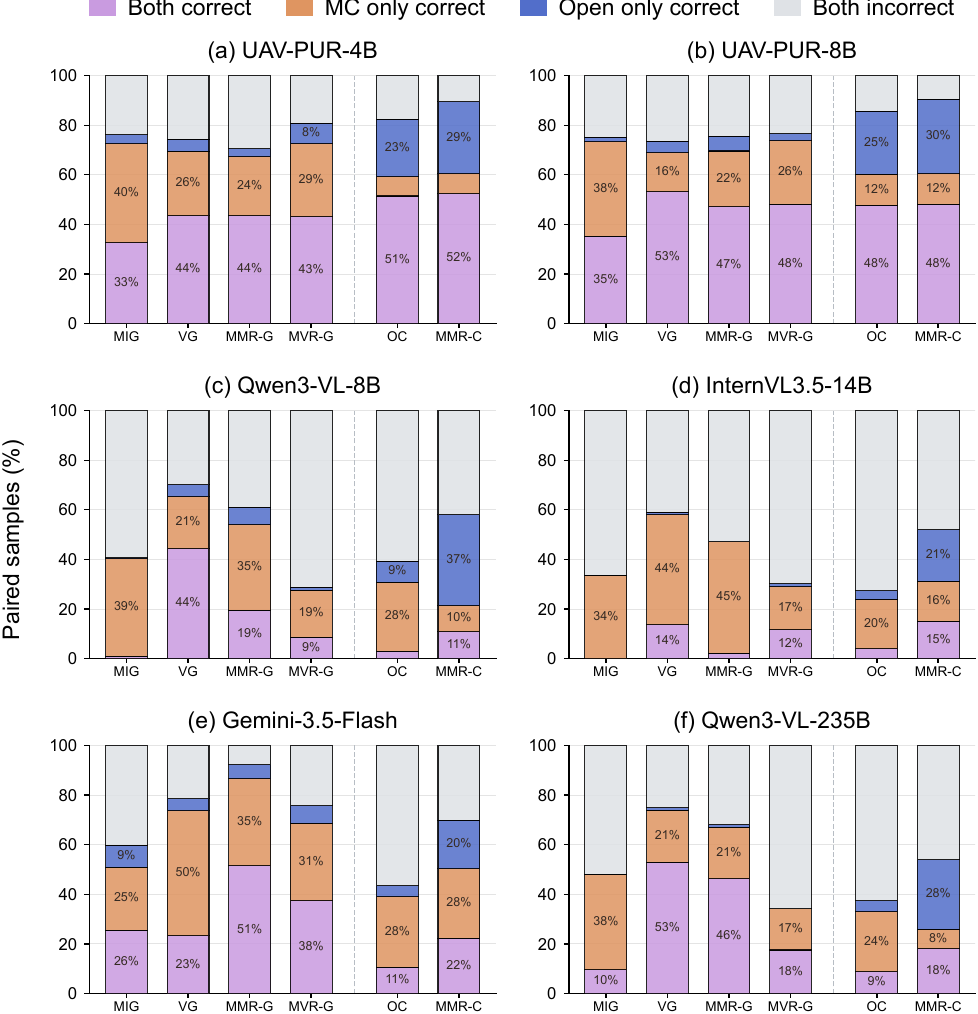}
\caption{Paired comparison of multiple-choice and open-ended predictions across six models. Stacked bars show the proportions of samples answered correctly in both formats, only one format, or neither.}
\label{fig_9}
\end{figure}

\noindent\textbf{Open-ended detection and grounding.}
Table~\ref{tab_4} reports performance when models predict bounding boxes directly from images and questions, without bounding-box options provided. Table~\ref{tab_4} includes models capable of directly generating valid bounding boxes.
We evaluate OD using Micro-F1 and the four grounding tasks using Acc@0.5. 
Accuracy drops substantially compared with multiple-choice evaluation, as open-ended grounding requires models to discover the target and reject more distractor regions without candidate boxes.
General-purpose models perform unevenly across tasks: Qwen3-VL-32B achieves the highest OD score, Qwen3-VL-235B leads on VG, and Gemini-3.5-flash performs best on other tasks. UAV-PUR-8B reaches 62.57\% on OD and 50.85\% on MVR-G, compared with 11.20\% and 9.75\% for Qwen3-VL-8B. It also outperforms Qwen-\textit{sft}-8B on all five tasks, most notably on OD. 

\noindent\textbf{Open-ended counting.}
Table~\ref{tab_5} reports MAE and MRA for OC and MMR-C, together with Avg. Under, the average of their underestimation rates over open-ended predictions. General-purpose models frequently underestimate, with Avg. Under ranging from 67\% to 90\%. 
These results reveal a systematic underestimation tendency in UAV object counting.
UAV-PUR-4B achieves the best MAE and MRA on both tasks: compared with the best general-purpose results, the MRA of OC rises from 29.01\% to 71.01\%, and the MAE of MMR-C falls from 6.03 to 3.98. It also reduces Avg. Under from 74.15\% for Qwen-\textit{sft}-4B to 54.46\%. 

\begin{figure*}[tb]
\centering
\includegraphics[width=0.96\linewidth]{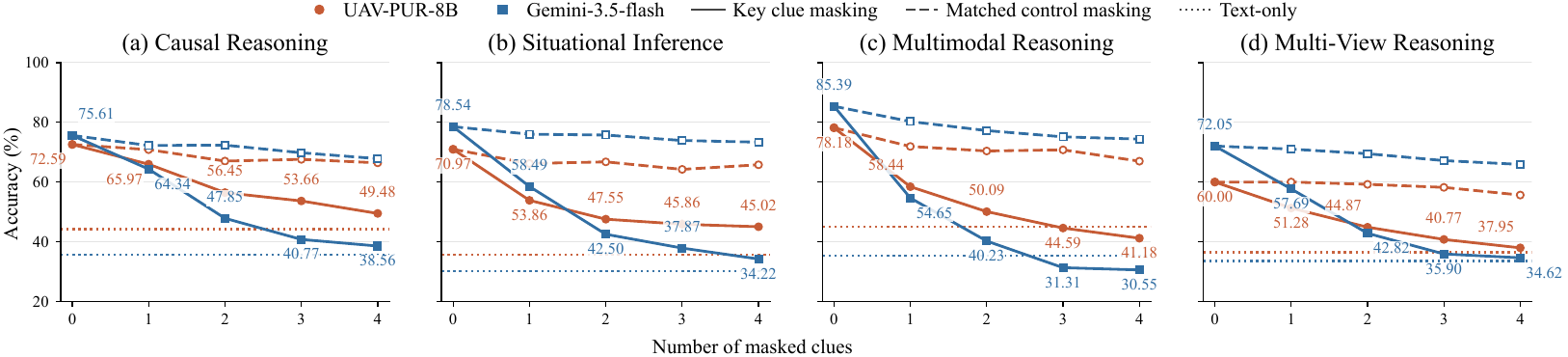}
\caption{Effects of visual clue masking on reasoning accuracy. On the same task subsets, progressive key-clue masking is compared with masking regions matched in area and shape. Zero denotes full input. Horizontal dotted lines indicate each model's text-only baseline.}
\label{fig_10}
\end{figure*}

\begin{figure*}[tb]
\centering
\includegraphics[width=0.96\linewidth]{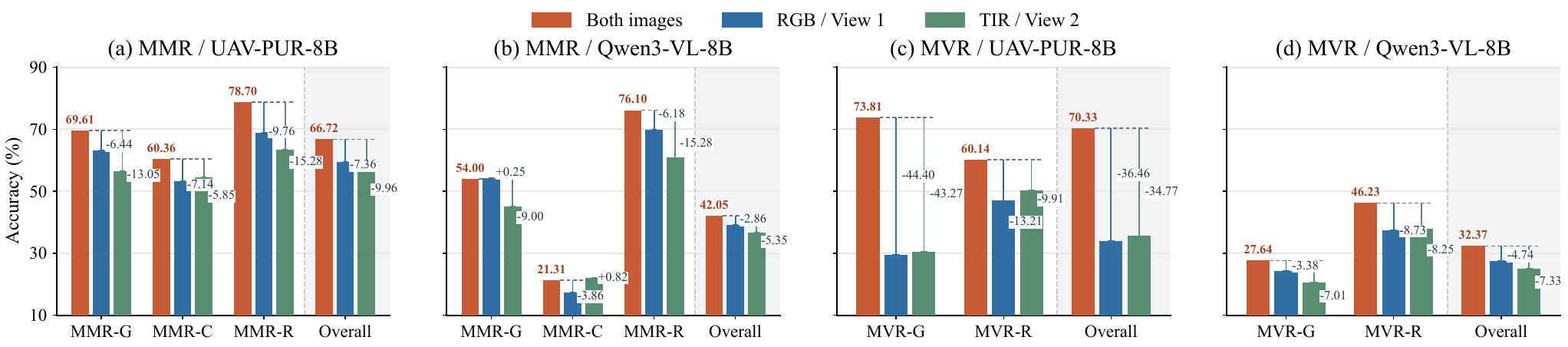}
\caption{Modality and view ablation study on MMR and MVR. Signed annotations indicate single-image accuracy changes relative to full-input performance.}
\label{fig_11}
\end{figure*}

\noindent\textbf{Paired comparison of answer formats.}
We evaluate six models on paired multiple-choice and open-ended versions. Each pair uses the same question, with candidate options provided for multiple-choice answering and omitted for direct box or count prediction. 
Open-ended grounding is evaluated using Acc@0.5, 
while counting predictions are considered correct when the relative error is below 25\%.
Fig.~\ref{fig_9} reports the proportions of paired samples for which both formats are correct, only multiple-choice answering is correct, only open-ended prediction is correct, or both are incorrect.  
Across grounding tasks, multiple-choice-only success consistently exceeds open-ended-only success, highlighting the greater difficulty of direct grounding compared with candidate box selection. Open-ended-only successes (blue segments in Fig.~\ref{fig_9}) still occur, but account for less than 10\% of samples in every grounding task—for example, 2.74\% on MVR-G for UAV-PUR-8B.
Compared with Qwen3-VL-8B, UAV-PUR-8B increases the proportion of samples answered correctly in both formats and reduces joint failures across all six tasks, with particularly pronounced improvements in MIG, cross-view grounding, and counting. 
Evaluating models in both formats provides a more comprehensive assessment of their capabilities.

\subsection{Ablation Study}


\noindent\textbf{Visual clue intervention.} 
We evaluate UAV-PUR-8B and Gemini-3.5-flash on subsets of CR, SI, MMR-R, and MVR-R containing 861, 713, 527, and 390 samples, respectively. 
Each sample contains at least four individually localizable key clues, each occupying at most 2\% of the corresponding image area. 
For each sample, we generate four random clue orderings and progressively mask the first one to four clues with gray patches, masking all corresponding regions across modalities or views. 
Control masks cover noncritical regions with matched areas and shapes. 
We average correctness across orderings within each sample and then across samples within each task. 
As shown in Fig.~\ref{fig_10}, accuracy decreases progressively across all four tasks as more key clues are masked. Masking noncritical regions has a much smaller effect. On MMR-R, masking four key clues leaves UAV-PUR-8B and Gemini-3.5-flash 3.79 and 4.74 percentage points below their respective text-only baselines. This suggests that the remaining visual content may mislead inference when essential clues are missing, resulting in worse performance than answering from the question and options alone. These results indicate that effective reasoning on these subsets depends on integrating multiple essential visual clues from different image regions. Global scene information and broader visual context cannot fully compensate for the absence of these clues.


\noindent\textbf{Modality and view ablation.}
To quantify the performance gains or losses from combining modalities or views, we evaluate UAV-PUR-8B and Qwen3-VL-8B on MMR and MVR using both images or either image alone. 
Single-image inputs retain RGB or TIR for MMR and View~1 or View~2 for MVR. 
Fig.~\ref{fig_11} reports subtask and overall accuracies, with single-image accuracy changes measured relative to the full-input condition.
Relative to paired inputs, the better single-image condition reduces accuracy more for UAV-PUR-8B than for Qwen3-VL-8B: 7.36 versus 2.86 percentage points on MMR and 34.77 versus 4.74 on MVR. Note that the substantial single-view drop on MVR-G primarily reflects the removal of essential reference or target information.
These benefits also extend to reasoning: paired inputs improve UAV-PUR-8B accuracy by 9.76 points on MMR-R over RGB alone and by 9.91 points on MVR-R over View~2 alone.
However, additional inputs do not consistently improve the original Qwen3-VL-8B across subtasks. RGB-only input achieves 0.25 points higher accuracy than paired inputs on MMR-G, while TIR-only input performs 0.82 points better on MMR-C. 
These small reversals suggest that additional modalities may introduce interference when visual clues are not effectively integrated, offsetting the potential gains from complementary information. 

\begin{table}[!t]
\centering
\caption{Ablation of training stages, auxiliary supervision, and Merger adaptation.
Within each size group, the best and second-best scores are shown in bold and underlined, respectively.}
\label{tab_6}
\small
\renewcommand{\arraystretch}{0.96}
\begin{tabular*}{\columnwidth}{@{\extracolsep{\fill}}lclccc@{}}
\toprule
Size & 3 Stages & Configuration & Per. & Und. & Rea. \\
\midrule
\multirow{8}{*}{4B}
 & \multirow{4}{*}{$\pmb{\times}$} & Base             & 48.66 & 57.87 & 46.83 \\
 &                                 & Stage 1 only     & 70.44 & 58.94 & 51.94 \\
 &                                 & Stages 1+2       & 73.94 & 65.88 & 65.99 \\
 &                                 & Qwen-\textit{sft} & 72.18 & 65.25 & 66.60 \\
\cmidrule(lr){2-6}
 & \multirow{4}{*}{$\pmb{\checkmark}$} & w/o Aux.    & 72.31 & 65.58 & 65.10 \\
 &                                     & Frozen Mergers    & 73.59 & 65.73 & 65.30 \\
 &                                     & S3: MC only      & \underline{74.58} & \textbf{67.05} & \underline{67.11} \\
 &                                     & \textbf{UAV-PUR} & \textbf{76.41} & \underline{66.60} & \textbf{67.97} \\
\midrule
\multirow{8}{*}{8B}
 & \multirow{4}{*}{$\pmb{\times}$} & Base             & 53.84 & 58.50 & 46.87 \\
 &                                 & Stage 1 only     & 70.22 & 57.63 & 52.77 \\
 &                                 & Stages 1+2       & 74.87 & 65.52 & 66.33 \\
 &                                 & Qwen-\textit{sft} & 74.96 & 65.67 & 67.47 \\
\cmidrule(lr){2-6}
 & \multirow{4}{*}{$\pmb{\checkmark}$} & w/o Aux.    & 74.25 & 65.94 & 67.21 \\
 &                                     & Frozen Mergers    & \underline{75.68} & 65.53 & 66.77 \\
 &                                     & S3: MC only      & 75.07 & \textbf{67.19} & \underline{68.22} \\
 &                                     & \textbf{UAV-PUR} & \textbf{77.03} & \underline{67.10} & \textbf{68.92} \\
\bottomrule
\end{tabular*}
\end{table}

\noindent\textbf{Fine-tuning ablation.} Table~\ref{tab_6} compares training stages and model configurations, with multiple-choice scores weighted by the number of test samples within each dimension. 
Base denotes the original Qwen3-VL model. Stage 1 only and Stages 1+2 evaluate the model after the corresponding training stages. Qwen-\textit{sft} is jointly fine-tuned on the 11 Drone-PUR tasks using only cross-entropy loss on answer tokens, without staged training. The remaining ablations retain all three stages: w/o Aux. removes the auxiliary heads and losses from the first two stages; the Frozen Mergers variant freezes the main visual merger and three DeepStack mergers; and S3: MC only uses only multiple-choice samples in the final stage. UAV-PUR is the complete configuration, which retains both multiple-choice and open-ended samples in Stage 3.

Stage 1 primarily improves perception, yielding gains of 21.78 and 16.38 percentage points for the 4B and 8B models, respectively. Training exclusively on perception samples in this stage also improves reasoning scores by 5.11 and 5.90 points, respectively, while understanding improves slightly at 4B but declines slightly at 8B. Adding Stage 2 increases reasoning accuracy by a further 14.05 and 13.56 percentage points. Stage 3 further improves all three dimensions at both model sizes. Compared with Qwen-\textit{sft}, the complete UAV-PUR achieves higher scores in all three dimensions at both model sizes. 
Under the evaluated training setup, removing auxiliary supervision or freezing the visual mergers lowers the multiple-choice scores in the perception and reasoning groups at both model sizes.
Using only multiple-choice samples in Stage 3 yields slightly higher understanding scores. Retaining open-ended samples, however, improves perception by 1.83 and 1.96 points and reasoning by 0.86 and 0.70 points for the 4B and 8B models, respectively. 
These gains show that retaining open-ended training samples in Stage 3 benefits perception and reasoning even under multiple-choice evaluation.

\subsection{Qualitative Analysis}

\begin{figure}[tb]
\centering
\includegraphics[width=0.85\linewidth]{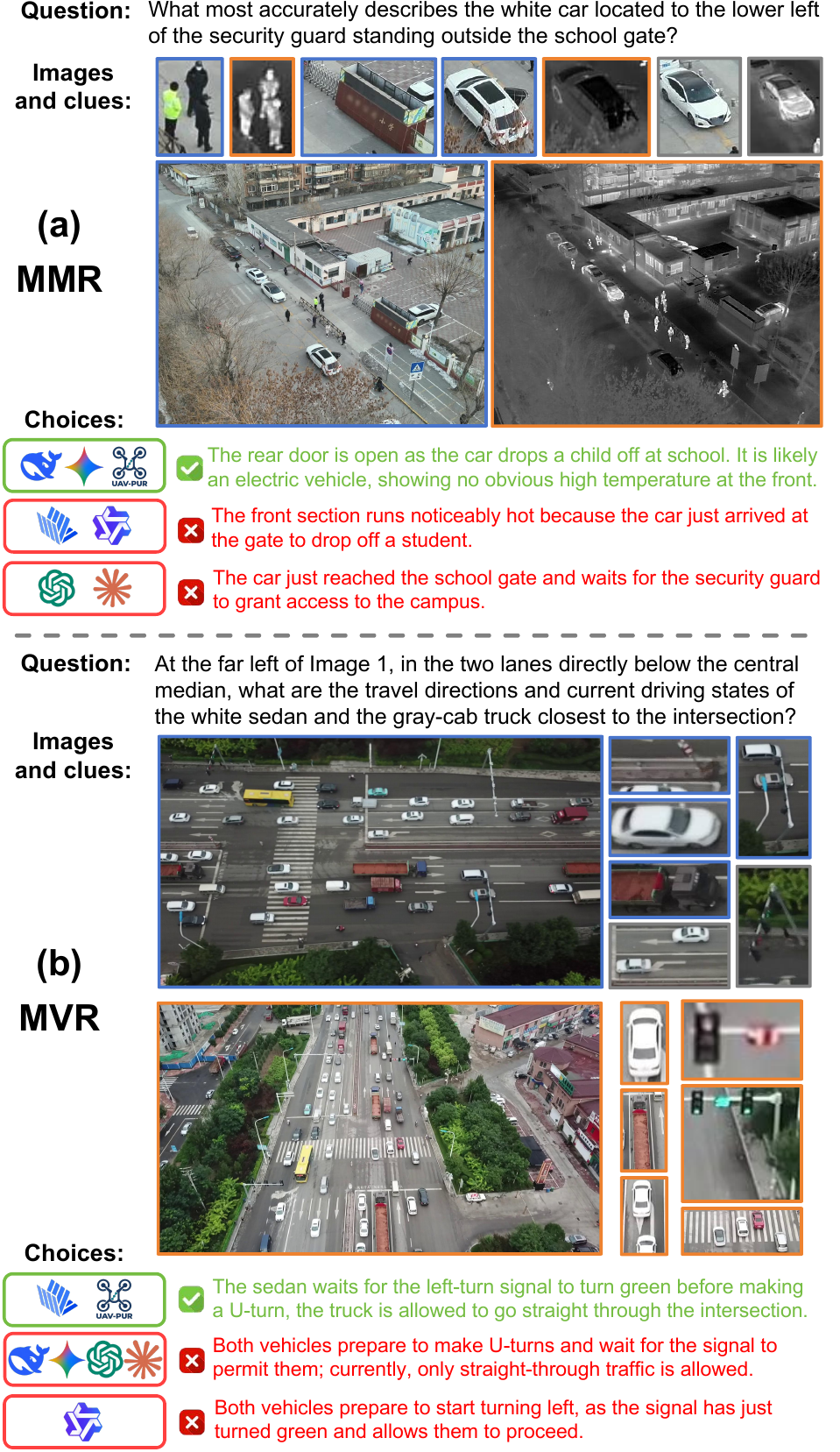}
\caption{Qualitative examples of multimodal and multi-view reasoning. Blue and orange borders mark essential clues from different input images; gray borders mark distracting regions. Model icons indicate selected options, with green and red denoting correct and incorrect answers, respectively.
Only the GT option and options selected by the displayed models are shown.
}
\label{fig_12}
\end{figure}


\noindent\textbf{Multimodal and multi-view reasoning.}
Fig.~\ref{fig_12} presents representative multiple-choice MMR and MVR examples, including questions, a subset of answer options, and enlarged clue regions.
Predictions are from several API-evaluated VLMs and our UAV-PUR-4B evaluated in Table~\ref{tab_1}. 
In the MMR example, RGB supports identification and localization of the guard and vehicle and reveals the open door, while TIR provides clues of the vehicle's thermal state. 
The MVR example requires locating two right-facing target vehicles in View~1 and associating them with the signals governing the upward-running lanes in View~2, which show green for straight travel and red for left turns. 
Road markings indicate a U-turn for the white car. 
Other vehicles in the truck's lane are proceeding straight through the intersection in both views, supporting the conclusion that the truck is permitted to proceed straight despite its current position behind the stop line. 
To reach a reliable answer, models must first establish spatial correspondences across inputs, locate the targets, and then identify and integrate essential clues while excluding distractors such as similar vehicles and unrelated markings.

\begin{figure*}[tb]
\centering
\includegraphics[width=0.91\linewidth]{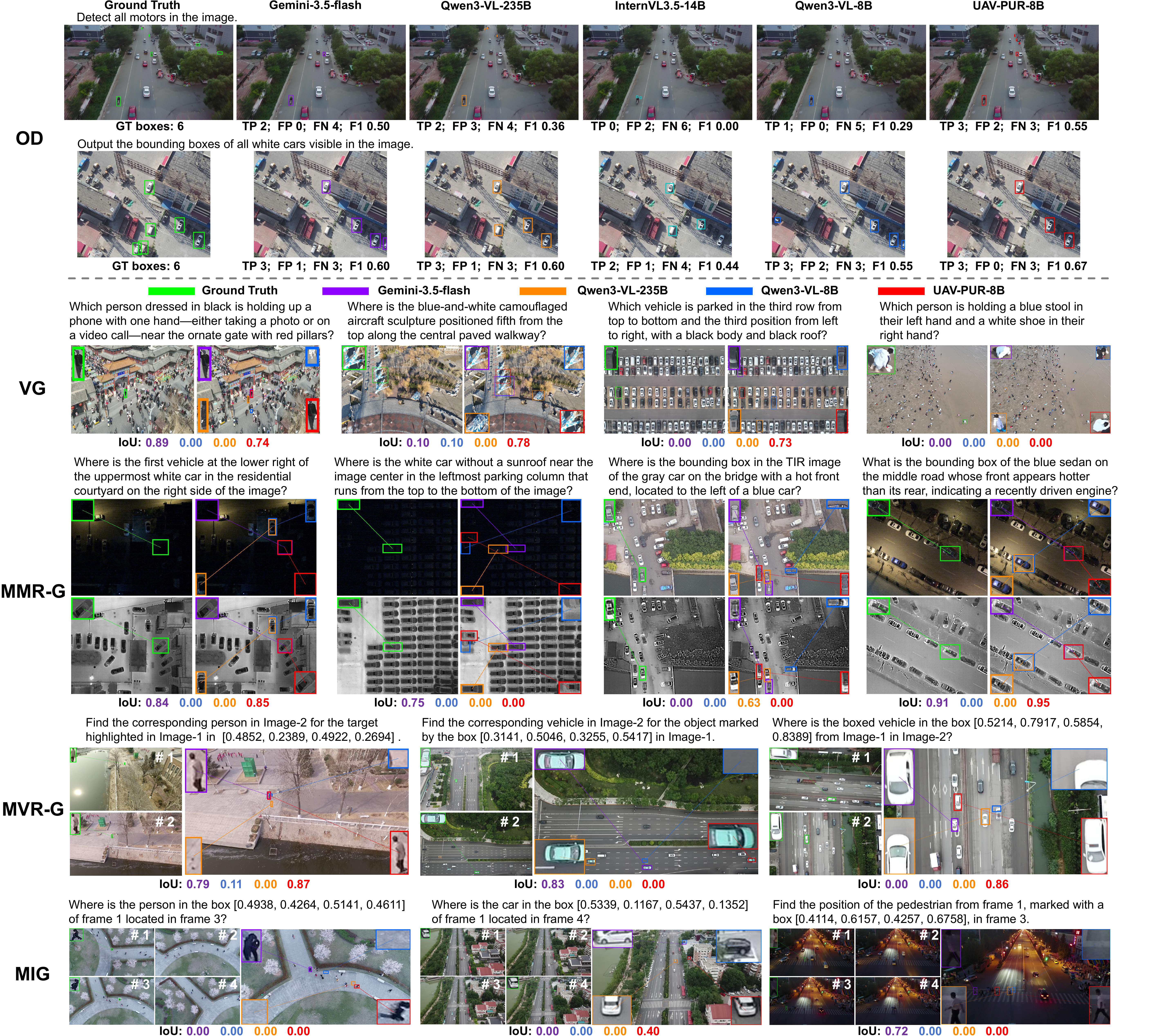}
\caption{Qualitative comparisons of open-ended detection and grounding on Drone-PUR. OD predictions are displayed separately for each model, with TP, FP, FN, and F1 scores reported below. 
For VG, MMR-G, MVR-G, and MIG, predictions from four models are overlaid on the target image, with corresponding IoU scores. Green boxes indicate ground truth; model predictions follow the color legend. Enlarged crops highlight target details and grounding differences.}
\label{fig_13}
\end{figure*}

\noindent\textbf{Open-ended detection and grounding.}
To further illustrate open-ended grounding errors, Fig.~\ref{fig_13} presents representative predictions across five tasks. 
In OD, missed detections of small objects remain common, and some models also produce false positives. 
All displayed models miss the even smaller motorcycles on the right side of the image. VG highlights difficulties in identifying instances through combinations of subtle attributes and spatial relations, including ordinal positions. 
Only UAV-PUR-8B correctly localizes the aircraft sculpture by ordinal position and the vehicle by parking row and column. 
However, all displayed models fail to identify the person described by the objects held in each hand, illustrating the difficulty of grounding combinations of fine-grained attributes.

MMR-G builds on VG by integrating complementary RGB and TIR clues. RGB provides color and appearance cues, while TIR reveals contours in low light and provides cues about material differences and thermal states. Gemini and UAV-PUR-8B correctly ground more targets in the displayed examples.
In MVR-G, Qwen3-VL-8B and Qwen3-VL-235B exhibit box offsets around candidate objects. 
Incorrect target identity is another source of error in MVR-G. Qwen3-VL-235B and UAV-PUR-8B select other cyan taxis in the second example, while Gemini and both Qwen models select other white cars in the third example. 
These errors suggest that models rely too heavily on appearance similarity and struggle to use spatial relations and scene context to distinguish targets across views.
Similar ambiguity occurs in MIG, where several incorrect predictions select people whose clothing resembles that of the target: black clothing in the first example and a white top with black trousers in the third.
Accurate identification across frames therefore requires spatial and temporal context beyond appearance similarity.
Overall, these examples highlight persistent challenges in semantic grounding and target correspondence across modalities, views, and frames. Reliable grounding requires sensitivity to subtle visual details and reasoning over spatial and temporal relationships.

\section{Conclusion}

In this paper, we introduced Drone-PUR, a dataset and benchmark that organizes 11 UAV tasks into perception, understanding, and reasoning according to how visual clues are discovered, interpreted, and connected. 
We also developed UAV-PUR, a baseline combining staged fine-tuning with auxiliary supervision, which improves performance across all three capability dimensions at both the 4B and 8B scales.
Evaluation of general-purpose VLMs reveals that strong object recognition can coexist with weak spatial judgment and uneven performance in multimodal and multi-view reasoning. 
Perception and reasoning performance are strongly correlated across the evaluated models. 
Single-task fine-tuning further reveals generally positive bidirectional transfer, with gains depending on the specific source and target tasks. 
Ablations show that reasoning depends on multiple essential visual clues whose absence cannot be fully compensated for by broader scene context. 
Open-ended grounding tasks further reveal difficulties in distinguishing similar targets. 
Despite substantial gains from domain adaptation, accurate spatial interpretation and reliable clue integration remain challenging.

\bibliographystyle{IEEEtran}   
\bibliography{reference}

\end{document}